\documentclass[11pt,a4paper]{article}

\usepackage[utf8]{inputenc}
\usepackage{amssymb}
\usepackage{authblk}
\usepackage{fullpage}
\usepackage{amsthm}
\usepackage{enumerate}
\usepackage{epsfig}
\usepackage{graphicx}
\usepackage{color}
\usepackage{amsmath}
\usepackage{hyperref}
\usepackage[mathscr]{euscript}
\usepackage{makeidx}

\usepackage{amsmath}
\usepackage{bm}
\usepackage{bbm}
\usepackage{amsfonts}
\usepackage{amssymb,amsthm}
\usepackage{graphicx}
\usepackage{algorithm}
\usepackage{algorithmic}

\newcommand{\bw}{\mathbf{w}}

\newcommand{\bx}{\mathbf{x}}
\newcommand{\bz}{\mathbf{z}}

\newcommand{\bfz}{\mathbf{z}}

\newcommand{\bmu}{\boldsymbol{\mu}}

\newcommand{\btheta}{\boldsymbol{\theta}}
\newcommand{\bepsilon}{\boldsymbol{\epsilon}}

\theoremstyle{remark}

\providecommand{\theoremname}{Theorem}
\providecommand{\propositionname}{Proposition}

  \theoremstyle{plain}

\author{Michalis K. Titsias \\
Department of Informatics, \\
Athens University of Economics and Business, \\
Athens, Greece
}

\begin{document}

\title{\bf Learning Model Reparametrizations: Implicit Variational Inference by Fitting  MCMC distributions}

\maketitle

\begin{abstract}
We introduce a new algorithm for approximate inference that combines reparametrization, 
Markov chain Monte Carlo and variational methods. We construct a very flexible implicit variational distribution 
synthesized by an arbitrary  Markov chain Monte Carlo  operation and a deterministic  
transformation that can be optimized using the reparametrization
trick. Unlike current methods for implicit 
variational inference, our method avoids the computation of log density ratios and therefore it is 
easily applicable to arbitrary continuous and differentiable models. We demonstrate the proposed 
algorithm for fitting banana-shaped distributions and for training variational autoencoders.   
\end{abstract}

\section{Introduction} 

Consider a probabilistic model with joint distribution   
$p(\bx,\bz)$ where $\bx$ are data and  $\bz$ are latent variables and/or random parameters.  
Suppose that exact inference in $p(\bx,\bz)$ is intractable which means that the posterior distribution
\begin{equation}
p(\bz | \bx) = \frac{p(\bx,\bz)}{\int p(\bx,\bz) d\bz}, \nonumber 
\end{equation}
is  difficult to compute due to the normalizing constant $p(\bx) = \int p(\bx,\bz) d\bz$
that represents the probability of the data and it is known as evidence or marginal likelihood.
The marginal likelihood is essential for estimation of any extra parameters in $p(\bx)$ or for model 
comparison. Approximate inference algorithms target to approximate $p(\bz | \bx)$ 
and/or $p(\bx)$. Two general frameworks, that we briefly review next,  are based on Markov chain Monte Carlo (MCMC) \cite{Robert2005,mcmchandbook2011}
and variational inference (VI) \cite{Jordan:1999, WainwrightJordan2008}.  

MCMC constructs a Markov chain with transition kernel 
$q(\bz'|\bz)$ that leaves $p(\bz|\bx)$ invariant, i.e.\ 
$$
p(\bz'|\bx) = \int q(\bz'|\bz) p(\bz|\bx) d \bz. 
$$ 
There exist many popular MCMC algorithms, such as random walk Metropolis-Hastings, Gibbs sampling, Metropolis-adjusted Langevin and 
Hamiltonian Monte Carlo; for detailed discussions see \cite{Robert2005,mcmchandbook2011}. The great advantage of MCMC is that it offers a very 
flexible non-parametric inference procedure that is  exact in the limit of long/infinite runs. 
The main disadvantage of MCMC is that it may require a long time to converge,  i.e.\ 
for the corresponding chain to produce an independent sample from $p(\bz|\bx)$.  Even after convergence 
the produced samples are statistically dependent and can exhibit high autocorrelation. 
Advances on adaptive MCMC \cite{haario-saksman-tamminen-01, andrieu-thoms-08}, Hamiltonian Monte Carlo 
\cite{duane-kennedy-pendleton-roweth-87, Neal2010MCMC, Hoffman2014} and Riemann  manifold methods \cite{Girolami11}
try to speed up MCMC convergence and improve on statistical efficiency, but the general problem is still far from been solved.    



In contrast, VI carries out approximate inference by fitting parametric distributions to the exact 
 distributions using optimization. To approximate the exact posterior 
$p(\bz|\bx)$ a distribution $q(\bz; \btheta)$ is introduced, that belongs to a 
tractable variational family, and the parameters $\btheta$ are fitted by minimizing a distance measure 
between the two distributions. The most widely used VI algorithms are based on   
minimizing the KL divergence  between $q(\bz; \btheta)$ and $p(\bz|\bx)$ so that  
$$
\btheta^* = \arg \min_{\btheta} \text{KL}(q(\bz;\btheta) || p(\bz | \bx)).
$$
This minimization can be equivalently expressed as 
a maximization of a lower bound on the log marginal likelihood, 
\begin{equation}
\mathcal{F}(\btheta) = \mathbbm{E}_{q(\bz;\btheta)}[\log p(\bx,\bz)-\log q(\bz;\btheta)] \leq \log p(\bx).
\label{eq:lowerIntro}
\end{equation}
An advantage of VI is that turns approximate inference into optimization 
which typically results in much faster algorithms than MCMC methods.  
The disadvantage, however, is that in order for the optimization procedure to be fast or tractable  
the parametric distribution $q(\bz;\btheta)$ needs to be oversimplified which often leads to  
poor approximations.    
An influential recent technique used in VI is stochastic optimization \cite{robbinsmonro51} that allows to extend 
traditional VI algorithms to big data \cite{HoffmanBB10,Hoffmanetal13} and to nonconjugate probabilistic models  
\cite{Carbonetto2009,Paisley2012,Ranganath2014, Mnih2014, Titsias2014_doubly, Salimans2013, Kingma2014, Rezende2014}.
Indeed, by using stochastic optimization VI has become a fully general inference method 
applicable to arbitrary probabilistic models, as those expressible by probabilistic programming
languages such as Stan \cite{stan:2017,Kucukelbir2016}.      
Stochastic optimization has also opened new directions towards increasing the 
flexibility of the parametric variational family by using normalizing flows \cite{RezendeM15} or by allowing 
$q(\bz;\btheta)$ to be an implicit distribution \cite{Ranganath16, SteinVariational16, WangLiu16, Karaletsos16, MeschederNG17, Huszar17, TranRB17}.
However, the increased flexibility of the parametric distribution $q(\bz;\btheta)$ often results in 
slow and less stable optimization.   
For instance, recent implicit variational methods 
\cite{Karaletsos16, MeschederNG17, Huszar17, TranRB17} that utilise arbitrarily flexible parametric families for 
 $q(\bz;\btheta)$ require the computation of log density ratios \cite{Sugiyama:2012}, which is a very difficult problem in high dimensional settings.
This implies that there is a (rather obvious) trade off between 
optimization efficiency and complexity of the parametric variational family; the more 
parameters and non-linear structure  $q(\bz;\btheta)$ has, the more difficult the optimization becomes. 

In this paper we propose to increase the flexibility of the variational family, and thus the approximation capacity of the inference procedure, not by adding more parameters and non-linearities but instead through the synthesis of MCMC and a model-based learnable reparametrization. On one hand, MCMC will allow us to increase the flexibility of the variational family by making it non-parametric and on the other hand the learnable 
reparametrization  will allows us to optimize the MCMC-based variational family by tuning a small set of parameters, such as parameters that define an affine transformation.    
This yields a MCMC-based implicit variational approximation that can be optimized using a procedure that, in terms of stability, 
behaves similarly to the standard reparametrization-based stochastic VI algorithms that fit Gaussian 
approximations \cite{Titsias2014_doubly,Kingma2014, Rezende2014}. Furthermore, unlike 
other implicit variational inference methods \cite{Karaletsos16, MeschederNG17, Huszar17, TranRB17}, the 
proposed algorithm does not require the computation of log density ratios.

The remainder of this paper has as follows.  Section \ref{sec:theory} develops the theory of the proposed method.
Section \ref{sec:related} discusses  related work such as the recent work in \cite{LiTL17}  that learns how to distil MCMC. 
Section \ref{sec:experiments}  presents experiments  on fitting banana-shaped distributions and for training variational autoencoders, while Section \ref{sec:discussion} 
concludes with a discussion. 


 
\section{Combine variational inference, MCMC and reparametrization}
\label{sec:theory}

Section \ref{sec:modelReparam} discusses a key concept of our method that is related to learnable 
model-based reparametrizations. Then Section  \ref{sec:varMCMC} introduces the algorithm 
that combines reparametrization,  MCMC and variational  inference.

\subsection{Model-based reparametrizations \label{sec:modelReparam}} 

The reparametrization trick plays a prominent role in modern 
approximate inference methods. The currently popular view of this 
trick is {\em approximation-based} where the reparametrization 
is assumed to be a characteristic of a variational family of distributions 
$\{q(\bz;\btheta), \forall \btheta \in \Theta\}$ so that a sample $\bz \sim q(\bz;\btheta)$ can be constructed by 
first drawing a latent noise variable $\bepsilon \sim p(\bepsilon)$
and then applying a transformation $\bz = g(\bepsilon; \btheta)$. 
Whether the distribution $q(\bz;\btheta)$ will be known explicitly or implicitly  
(i.e.\ its density will be analytic or not) depends on the properties of both the noise
distribution $p(\bepsilon)$ and the mapping $g(\bepsilon; \btheta)$. 
Approximate inference then proceeds by fitting  
$q(\bz;\btheta)$ to the exact posterior distribution. This is typically 
based on the standard procedure that minimizes the KL divergence $\text{KL}(q(\bz;\btheta)||p(\bz|\bx))$. 
The ability to reparametrize $q(\bz;\btheta)$ allows us to speed up the 
stochastic optimization of the variational parameters $\btheta$ by making use of 
gradients of the log model density $\log p(\bx,\bz)$ \cite{Titsias2014_doubly,Kingma2014, Rezende2014, Ruiz2016}. 
To summarize in the approximation-based view of reparametrization we leave the model 
$p(\bx, \bz)$ unchanged and we reparametrize some 
variational approximation to this model. 

An alternative way to apply reparametrization  is {\em model-based}. Here, 
we reparametrize directly the model $p(\bx, \bz)$ in order to obtain a new model representation and then we 
apply (approximate) inference in this new representation. The inference procedure could be based on any 
 framework such as exact inference (if possible), MCMC or variational methods. 
The objectives of model-based reparametrizations are similar with the ones of approximation-based 
reparametrizations, i.e.\ we would like to make inference and parameter estimation as easy
as possible. Several examples of  model-based reparametrizations have been used in the literature, such as the 
centered and non-centered reparametrizations \cite{ncp}.    
These reparametrizations are typically 
model-specific and they are not learnable during inference. Next we develop a generic 
methodology  
for continuous spaces and in Section \ref{sec:varMCMC}  we develop an algorithm 
that automatically can learn model-based reparametrizations.


A valid repametrization of  $p(\bx, \bz)$ essentially performs
a transformation of $\bz$ into a new set of random variables $\bepsilon$ so that  
the probability of the observed data $p(\bx)$ remains unchanged, i.e.\ we have 
the following marginal likelihood invariance property
$$
p(\bx) = \int p(\bx, \bz) d \bz = \int p(\bx, \bepsilon) d \bepsilon.
$$
Given that $\bz$ is continuous and $p(\bx, \bz)$ is differentiable w.r.t.\ $\bz$ a
generic way to define a valid reparametrization is through an invertible transformation
$$
\bepsilon = g^{-1}(\bz;\btheta), \ \ \ \bz = g(\bepsilon;\btheta). 
$$ 
Then the reparametrized model becomes
\begin{equation}
p(\bx,\bepsilon; \btheta) = p(\bx, g(\bepsilon;\btheta)) J(\bepsilon;  \btheta),
\label{eq:reparamModel}
\end{equation}
where  $J(\bepsilon;  \btheta)  = | \text{det} \nabla_{\bepsilon} g(\bepsilon; \btheta) |$  
denotes the determinant of the Jacobian of the transformation. The marginal likelihood is 
$$
p(\bx) =  \int p(\bx, g(\bepsilon;\btheta)) J(\bepsilon;  \btheta) d \bepsilon, 
$$
which clearly is equal to the marginal likelihood $\int p(\bx, \bz) d \bz$ computed from the initial representation.
Thus, a first noticeable property of the reparametrized model is that while  $\btheta$  
enters the expression  in  \eqref{eq:reparamModel} superficially as a {\em new model parameter}, it does not 
change the value of the marginal probability $p(\bx)$.
Therefore, $\btheta$ is not a model parameter that can influence the fit to the observed data $\bx$. 
In contrast the reparametrized posterior distribution 
$p(\bepsilon|\bx,\btheta) \propto  p(\bx, g(\bepsilon;\btheta)) J(\bepsilon; \btheta)$  
does depend on $\btheta$,  although notice that by transforming an exact sample from $p(\bepsilon|\bx,\btheta)$ 
back to the space of $\bz$ we always get an exact sample from $p(\bz|\bx)$ irrespectively of the value of $\btheta$. 
It is precisely this dependence of $p(\bepsilon|\bx,\btheta)$ on $\btheta$ that we would like 
to exploit in order to improve the accuracy and/or computational efficiency of approximate inference methods. 
For instance, if $g(\bepsilon;\btheta)$ is chosen so that $p(\bepsilon|\bx,\btheta)$ becomes roughly
independent across the dimensions of $\bepsilon$, then simple inference methods such mean 
field variational approximations or Gibbs sampling will be very effective.\footnote{The mean field 
approximation will capture accurately the dependence structure of the posterior, while Gibbs sampling 
will enjoy high statistical efficiency producing essentially an independent sample in every iteration.}     

Next we introduce a variational inference and MCMC hybrid algorithm that directly operates in the reparametrized model and it can 
auto-tune its performance by optimizing $\btheta$.  

\subsection{Implicit  MCMC-based variational inference \label{sec:varMCMC}}

Assume an arbitrary MCMC algorithm, such as standard random walk Metrololis-Hastings or more advances schemes like
 Hamiltonian Monte Carlo that is very popular in probabilistic programming systems \cite{stan:2017}, having a transition 
kernel $q(\bepsilon'|\bepsilon)$ that leaves the posterior distribution $p(\bepsilon|\bx, \btheta)$ invariant, i.e.\ 
$$
p(\bepsilon'|\bx, \btheta)  = \int q(\bepsilon'|\bepsilon) p(\bepsilon|\bx, \btheta) d \bepsilon.
$$ 
Given some specific value for $\btheta$, if we apply the MCMC algorithm for long enough we will eventually converge 
to the posterior distribution. This translates to convergence to the initial
posterior distribution $p(\bz|\bx)$ irrespectively of the chosen value $\btheta$. 
This somehow gives us the freedom to adapt $\btheta$ so that to speed up convergence to 
$p(\bepsilon|\bx, \btheta)$ and essentially to $p(\bz|\bx)$. So we need  
a learning procedure to tune $\btheta$ and next we shall use
variational inference. Specifically, we assume that MCMC is initialized with some 
fixed distribution $q_0(\bepsilon)$ and then the transition kernel is applied for
a fixed number of $t$ iterations. This constructs the marginal MCMC distribution 
$$
q_t(\bepsilon) = \int Q_t(\bepsilon|\bepsilon_0)  q_0(\bepsilon_0) d \bepsilon_0,
$$ 
where $Q_t(\bepsilon'|\bepsilon_0)$ is the compound transition density obtained 
by applying $t$ times the transition kernel $q(\bepsilon'|\bepsilon)$. $q_t(\bepsilon)$ is an implicit distribution 
since we can only draw samples from it but we cannot evaluate $q_t(\bepsilon)$. Nevertheless, 
by following the recent advances in implicit variational inference \cite{Karaletsos16, MeschederNG17, Huszar17, TranRB17}  
we shall consider 
$q_t(\bepsilon)$ as the variational distribution and fit it to the exact posterior $p(\bepsilon|\bx,\btheta)$ 
by minimizing the KL divergence $\text{KL}(q_t(\bepsilon) || p(\bepsilon|\bx,\btheta))$ which leads to the 
maximization of the standard lower bound 
\begin{equation}
\mathcal{F}_t(\btheta) = \int q_t(\bepsilon) \left[ 
 \log p(\bx, g(\bepsilon;\btheta)) J(\bepsilon;  \btheta)  - \log q_t(\bepsilon) \right] d \bepsilon. 
\label{eq:mcmcBound} 
\end{equation}    
Unlike the exact marginal likelihood, this bound does depend 
on the parameters $\btheta$ which now become variational parameters that can be tuned to improve 
the approximation. Also notice that, unlike the standard way of thinking about variational lower bounds where the variational parameters are defined to 
directly influence the variational 
distribution, in the above bound the variational parameters $\btheta$ are essentially part of the 
log joint density and rather indirectly influence $q_t(\bepsilon)$. Thus, when we maximize 
$\mathcal{F}_t(\btheta)$ we essentially try to transform the reparametrized exact posterior distribution 
$p(\bepsilon|\bx,\btheta)$ so that to make it more easily reachable by the MCMC marginal $q_t(\bepsilon)$. 
We can express the equivalent standard view of the variational lower bound in \eqref{eq:mcmcBound} 
by changing variables and re-writing $\mathcal{F}_t(\btheta)$ as
$$
\mathcal{F}_t(\btheta) = \int q_t(\bz; \btheta) \left[ \log p(\bx, \bz) 
- \log q_t(\bz; \btheta) \right] d \bz, 
$$
where 
$$
q_t(\bz; \btheta)  = q_t( g^{-1}(\bfz; \btheta) ) \widetilde{J} (\bz; \btheta)
$$
and $ \widetilde{J} (\bz;  \btheta)  = | \text{det} \nabla_{\bz} g^{-1}(\bz; \btheta) |$ denotes the determinant of the Jacobian 
associated with the inverse transformation $g^{-1}(\bz; \btheta)$. This clearly shows that 
 the induced variational distribution $q_t(\bz; \btheta)$ is reparametrizable and is made   up
by an implicit MCMC marginal $q_t(\bepsilon)$, that operates in the space of $\bepsilon$, 
followed by an invertible transformation that maps $\bepsilon$ to $\bz$. 

To maximize \eqref{eq:mcmcBound} w.r.t.\ $\btheta$ we can use an algorithm similar to 
Monte Carlo EM  \cite{Wei1990}. Suppose we are at the $k$-th iteration of the 
optimization, where the previous value of the parameter vector is $\btheta_{k-1}$. We
first perform an E-step where MCMC dynamics flow through the unnormalized posterior 
$p(\bx, g(\bepsilon;\btheta_{k-1})) J(\bepsilon;  \btheta_{k-1})$ for $t$ iterations which 
produces an independent sample from $q_t(\bepsilon)$. By repeating this process $S$ times we can produce 
a set of $S$ independent samples $\{\bepsilon^{(s)}\}_{s=1}^S$. 
Then we perform an M-step where we keep the approximate posterior $q_t(\bepsilon)$ fixed 
and we maximize the following Monte Carlo approximation of \eqref{eq:mcmcBound} with respect to $\btheta$
$$
\frac{1}{S} \sum_{s=1}^S \log p(\bx, g(\bepsilon^{(s)};\btheta)) J(\bepsilon^{(s)};  \btheta), 
$$     
where the constant entropic term $-\int q_t(\bepsilon) \log q_t(\bepsilon) d \bepsilon$ 
has been dropped since it does not depend on $\btheta$
($q_t(\bepsilon)$  depends on the old values $\btheta_{k-1}$). We can take a step towards maximizing this Monte Carlo 
objective by performing a gradient ascent step, 
$$
\btheta_{k} = \btheta_{k-1}  + \eta_k \frac{1}{S} \sum_{s=1}^S \nabla_{\btheta = \btheta_{k-1}} \log p(\bx, g(\bepsilon^{(s)}; \btheta )) J(\bepsilon^{(s)} ;  \btheta ), 
$$       
where $\eta_{k}$ denotes the learning rate. Algorithm 1 shows all steps of the procedure
where for simplicity we have assumed that $S=1$ and we have added another update for any possible 
model parameters $\bw$ that determine the joint probability density and for which we do point estimation.    

To implement the algorithm the user needs to specify: i) the invertible 
transformation $g(\bepsilon; \btheta)$, ii) the initial MCMC distribution $q_0(\bepsilon)$ and 
iii) the MCMC kernel density together with the number of iterations $t$ that overall define the implicit
density $Q_t(\bepsilon'|\bepsilon_0)$.  
By varying the above options we are truly blending between variational inference 
and MCMC and for instance we can easily recognise the following extremes cases: 

\begin{enumerate}

\item When $t=0$ (i.e.\ there is no MCMC and $q_t(\bepsilon)$ collapses to $q_0(\bepsilon)$), 
      $g(\bepsilon; \btheta) = L \bepsilon + \bmu$ and  $q_0(\bepsilon) = \mathcal{N}(\bepsilon|{\bf 0}, I)$
      the method becomes the standard reparametrization-based variational inference algorithm 
      that fits a Gaussian approximation to the true posterior \cite{Titsias2014_doubly, ChallisB11}.      

\item When $t=0$, $g(\bepsilon; \btheta) = \bepsilon + \bmu$ and  $q_0(\bepsilon)= \delta(\bepsilon)$
      the method leads to maximum a posteriori estimation. 

\item When $g(\bepsilon; \btheta) = \bepsilon$ and we consider large $t$, the method reduces to standard Monte Carlo
      EM or standard MCMC (in case there are no model parameters $\bw$ for which we do point estimation).     

\end{enumerate}
Furthermore, according to the well-known inequality $\text{KL}(q_t(\bepsilon) || p(\bepsilon|\bx,\btheta))
\leq \text{KL}(q_{t-1}(\bepsilon) || p(\bepsilon|\bx,\btheta))$ 
\cite{Cover:1991} the approximation monotonically improves as we increase the number $t$ of MCMC iterations. 
Specifically, from this inequality and the fact that the MCMC marginal $q_t(\bepsilon)$ converges
to  $p(\bepsilon|\bx,\btheta)$ as $t$ increases we can conclude that the lower bound in \eqref{eq:mcmcBound}  
satisfies  
$$
\mathcal{F}_t(\btheta) \geq \mathcal{F}_{t-1}(\btheta), \ \ \ \mathcal{F}_{\infty}(\btheta) = \log p(\bx).
$$  
While this monotonic  improvement with the iterations holds also for the regular MCMC, 
the promise
of our approach is that by learning the reparametrization parameters $\btheta$ we can obtain efficient 
approximations even with small $t$. An illustrative example of this will be given in Section \ref{sec:banana} using a banana-shaped 
distribution. 
          
 \begin{algorithm}[tb]
   \caption{Implicit MCMC-based variational inference using reparametrization \label{algorm}}
   \label{alg1}
\begin{algorithmic}
   \STATE Initialize $\btheta_0$, $\bw_0$, $k=0$
    \REPEAT 
    \STATE  k = k+1
    \STATE Update reparametrization parameters: 
      \STATE \ \ \ \ \ \% initialize MCMC:
      \STATE \ \ \ \ \ $ \bepsilon_0 \sim q_0(\bepsilon_0)$, 
      \STATE \ \ \ \ \ \% MCMC for $t$ iterations targeting the unnormalized $p(\bx, g(\bepsilon;\btheta_{k-1})) J(\bepsilon;  \btheta_{k-1})$: 
      \STATE \ \ \ \ \ $ \bepsilon \sim Q_t(\bepsilon | \bepsilon_0)$,  
         \STATE \ \ \ \ \ $    
         \btheta_k = \btheta_{k-1}  + \eta_k 
          \nabla_{\btheta = \btheta_{k-1} } \log p(\bx, g(\bepsilon; \btheta )) J(\bepsilon;  \btheta ), 
           $
    \STATE Update model parameters: 
     \STATE \ \ \ \ \ $ \bepsilon_0 \sim q_0(\bepsilon_0)$,  
       \STATE \ \ \ \ \ \% MCMC for $t$ iterations targeting the unnormalized $p(\bx, g(\bepsilon;\btheta_k)) J(\bepsilon;  \btheta_k)$: 
      \STATE \ \ \ \ \  $\bepsilon \sim Q_t(\bepsilon | \bepsilon_0)$, 
       \STATE \ \ \ \ \ $\bz = g(\bepsilon; \btheta_{k}),$
     \STATE \ \ \ \ \ $\bw_k = \bw_{k-1}  + \rho_k 
          \nabla_{\bw = \bw_{k-1}} \log p(\bx, \bz),$ 
    \UNTIL{convergence criterion is met.}  
\end{algorithmic}
\end{algorithm}

\section{Related work \label{sec:related}} 


The proposed algorithm performs approximate inference by fitting an implicit distribution 
and therefore it shares similarities with other implicit variational inference methods such as 
\cite{Ranganath16, SteinVariational16, WangLiu16, Karaletsos16, MeschederNG17, Huszar17, TranRB17}. 
Such approaches construct an implicit distribution by considering a tractable noise distribution $q(\bepsilon)$ 
and then passing $\bepsilon$ through a complex non-invertible neural network mapping $\bz = f(\bepsilon;\btheta)$. In contrast, 
our method does the opposite, i.e.\  it considers an implicit complex noise distribution  $q_t(\bepsilon)$ (obtained by MCMC)  and it passes $\bepsilon$ 
through a simple (invertible) transformation  $\bz = g(\bepsilon;\btheta)$. This avoids the need for computing density ratios 
which is a significant computational advantage of our approach relative to other implicit variational inference methods.   
 Furthermore, our method can be viewed as an extension of the standard reparametrization-based stochastic VI algorithms that fit Gaussian 
approximations \cite{Titsias2014_doubly,Kingma2014, Rezende2014}. 
Specifically,  \cite{Titsias2014_doubly} considers a reparametrization-based variational approximation having two separate 
components that the user can specify independently: the so-called neutral or noise distribution and a deterministic  transformation.  
Here, we allowed the neutral distribution in \cite{Titsias2014_doubly} to become 
an implicit MCMC distribution, while before it had a simple parametric form such as Gaussian or uniform.  

Regarding other MCMC and VI hybrid algorithms our method is closely related to the recent work in 
\cite{LiTL17} which is based on learning how to distil MCMC. Our method differs since it directly maximizes  
a single cost function which is the standard lower bound, while in \cite{LiTL17} 
there is an additional cost function that tries to project $q(\bz; \btheta)$ to the MCMC distribution. 
The use of two cost functions could cause convergence problems. 
Notice, however, that the approach in  \cite{LiTL17}  is fully general while our method is applicable only to 
continuous and differentiable probabilistic models.  A more distantly related method is the work in \cite{icml2015_salimans15} which combines MCMC and variational
inference by introducing several auxiliary variables (associated with the MCMC iterations) 
and corresponding auxiliary models that need to be fitted. Our method instead directly fits an MCMC distribution  to the 
exact posterior distribution without requiring the augmentation with  auxiliary variables  and learning auxiliary models.  
Finally, another more distantly related  method that  interpolates between Langevin dynamics \cite{WellingT11} and reparametrization-based 
stochastic VI was introduced in \cite{Domke17}.

\section{Experiments  \label{sec:experiments} }

Section \ref{sec:banana} considers a banana-shaped distribution, while Section 
\ref{sec:varautoencode} applies the proposed method to amortised inference 
and variational autoencoders.

\subsection{Fitting a banana-shaped distribution  \label{sec:banana}}

\begin{figure*}[!htb]
\centering
\begin{tabular}{cc}
{\includegraphics[scale=0.4]
{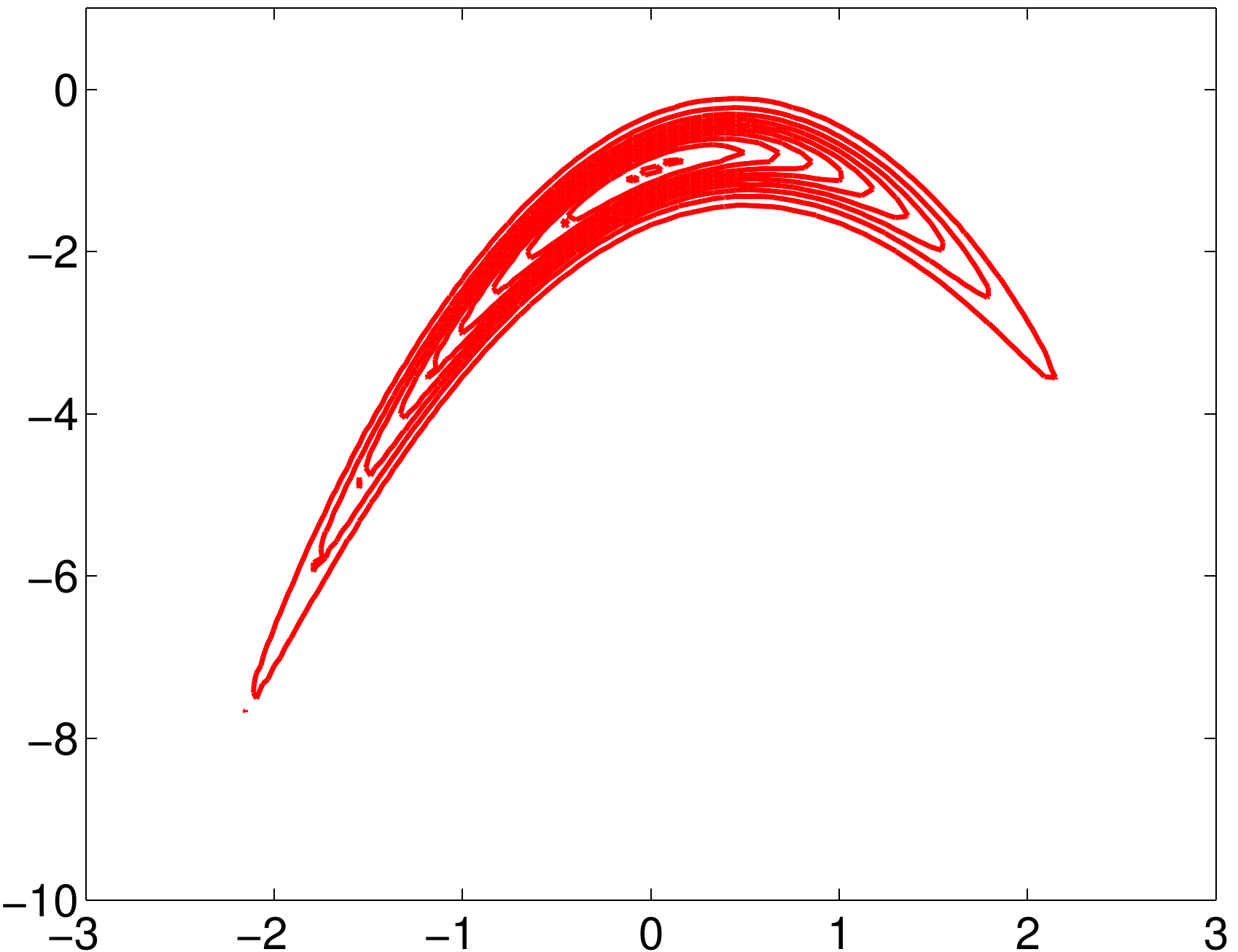}} &
{\includegraphics[scale=0.4]
{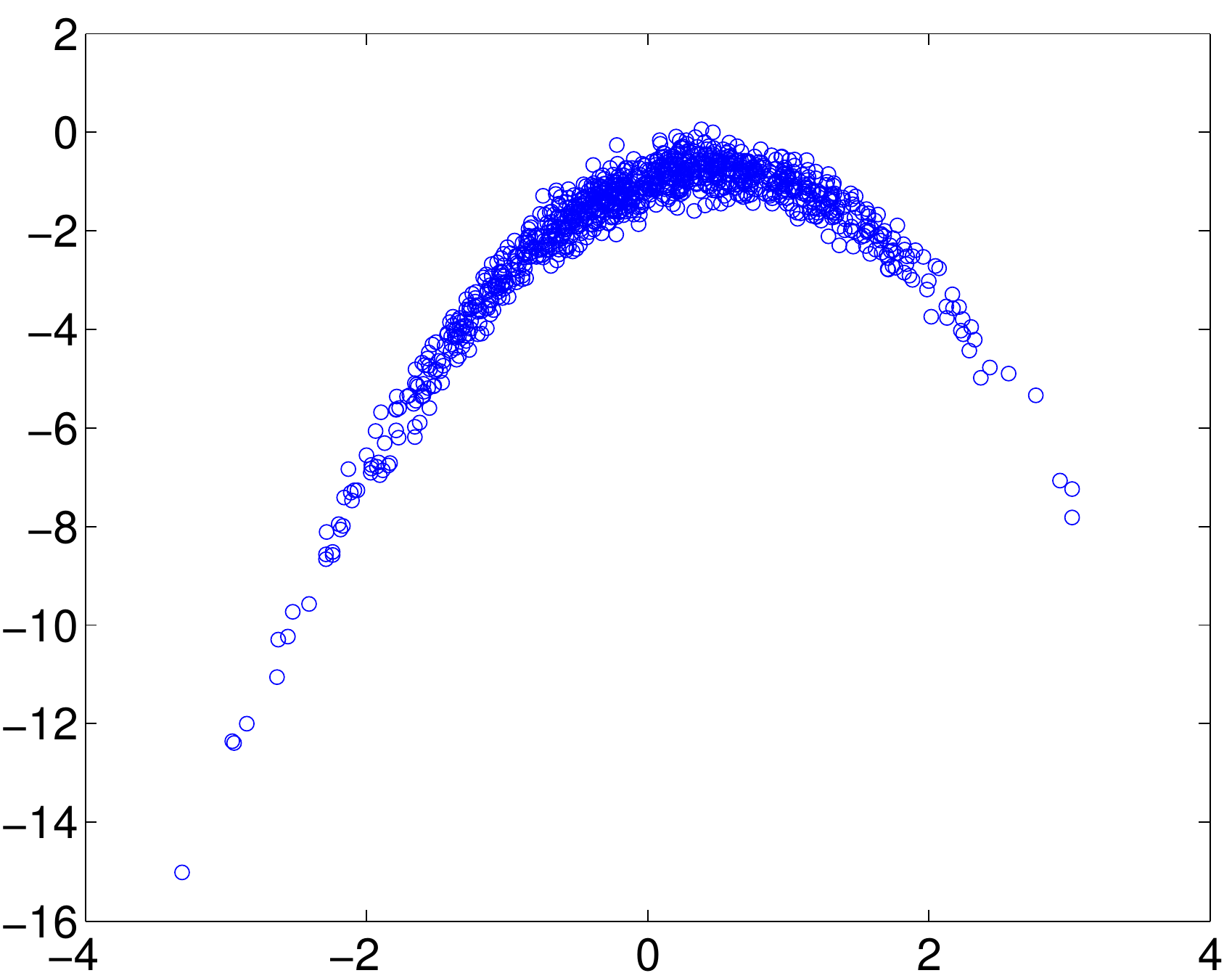}}  \\
(a)  & (b)
\end{tabular}
\caption{Panel (a) shows contours of the banana-shaped distribution, while panel (b) shows $1000$ independent samples.} 
\label{fig:bananaDistr}
\end{figure*}

\begin{figure*}[!htb]
\centering
\begin{tabular}{ccc}
{\includegraphics[scale=0.28]
{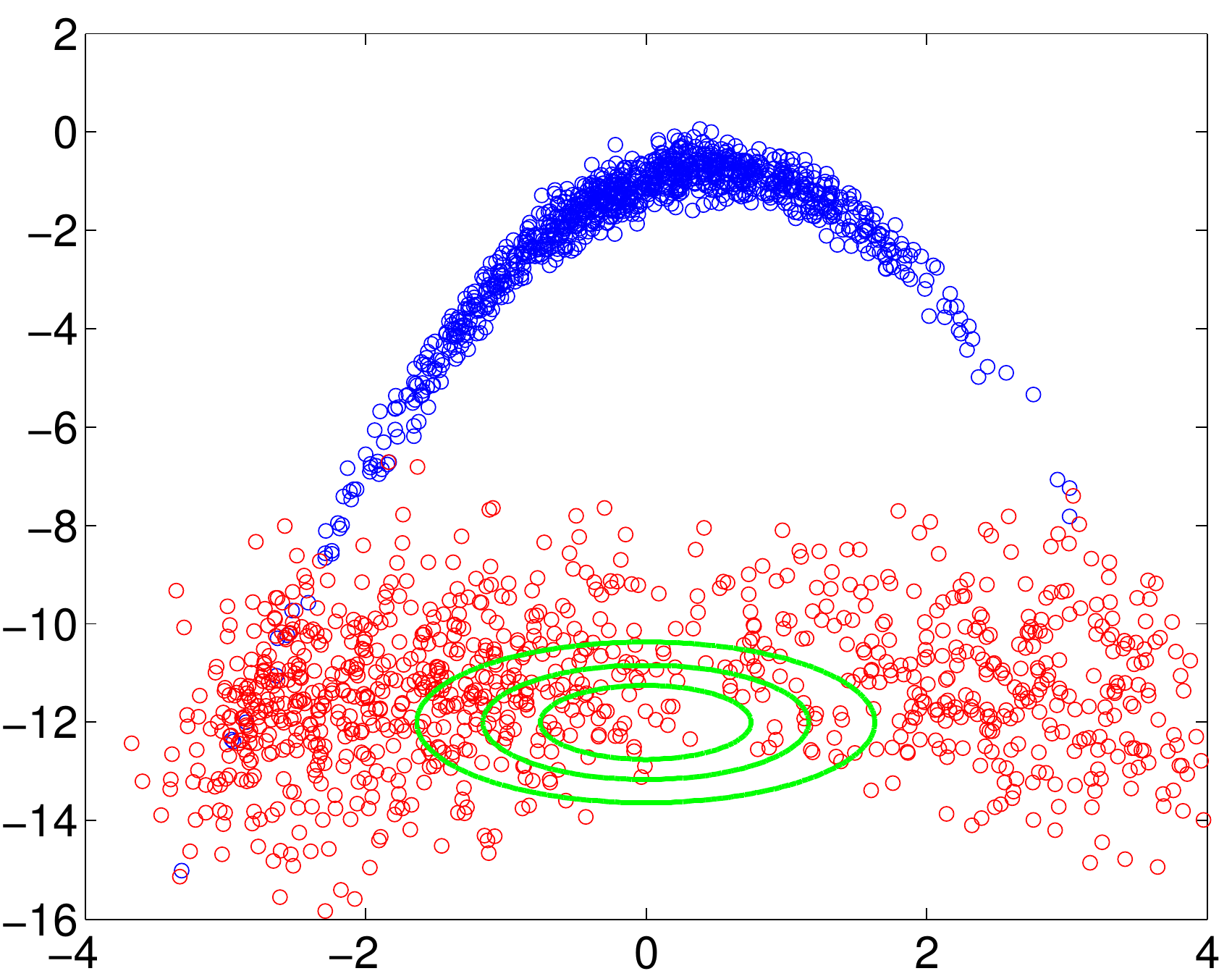}} &
{\includegraphics[scale=0.28]
{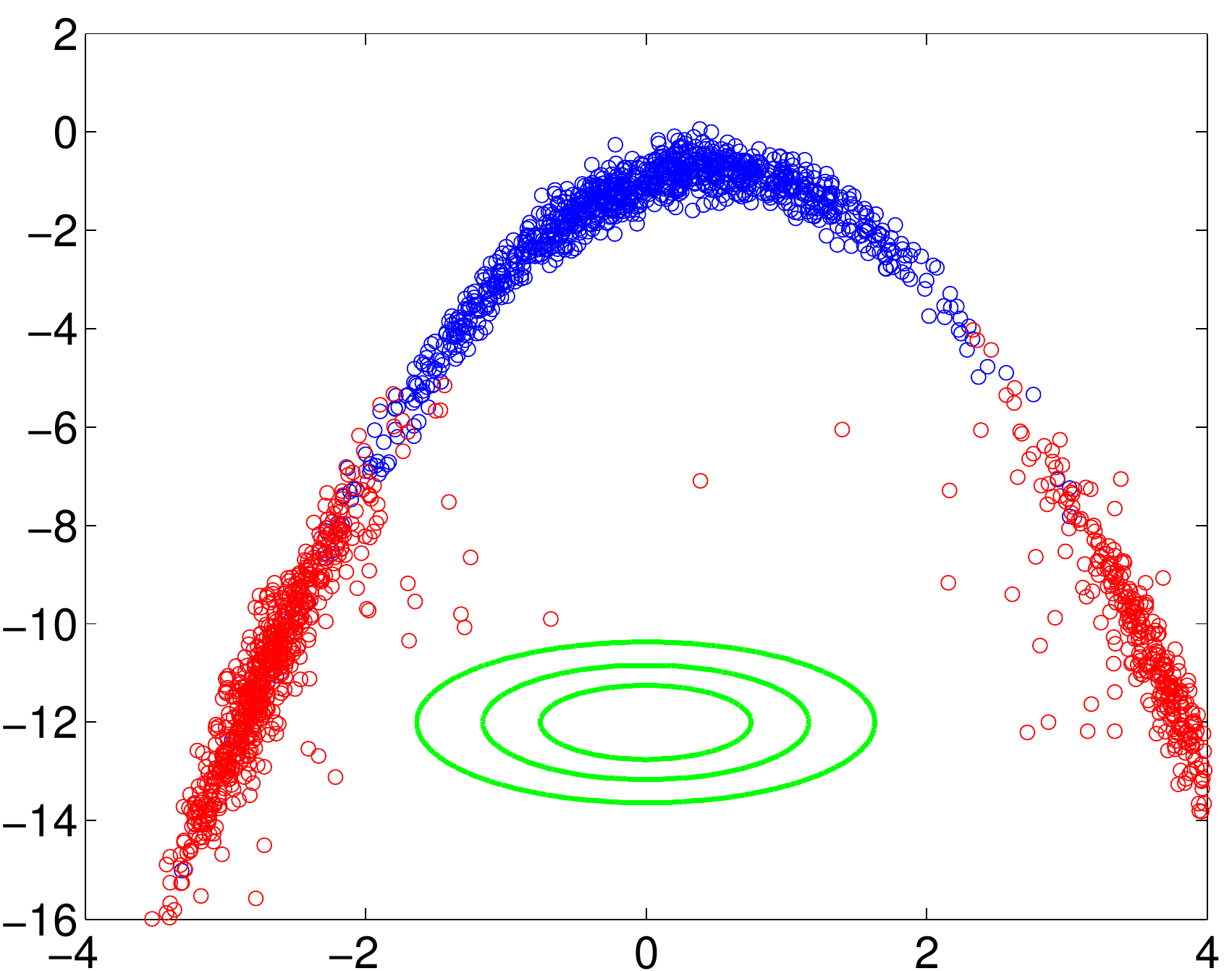}} &
{\includegraphics[scale=0.28]  
{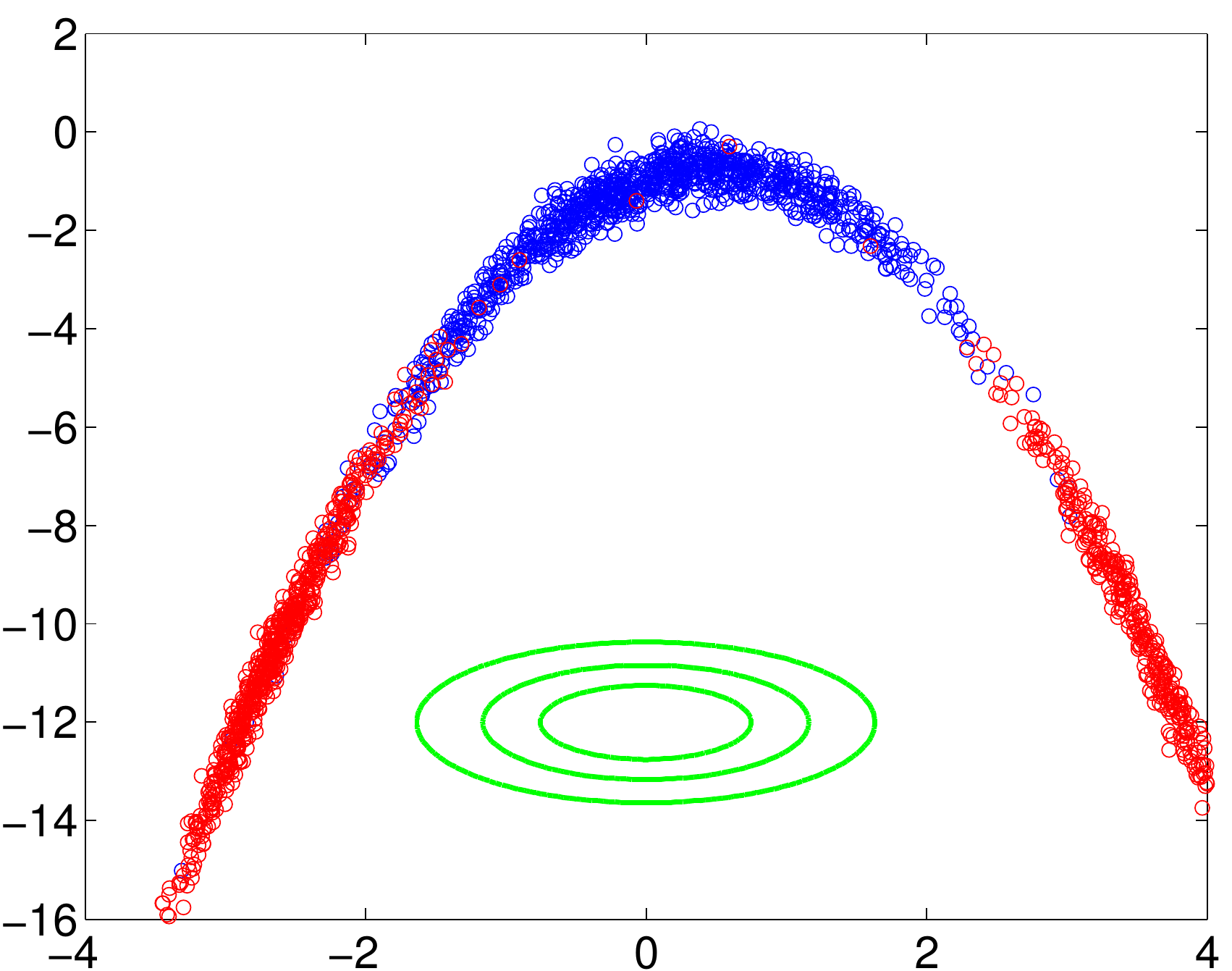}} \\
$t=5$ & $t=20$ &  $t=50$ \\ 
{\includegraphics[scale=0.28]  
{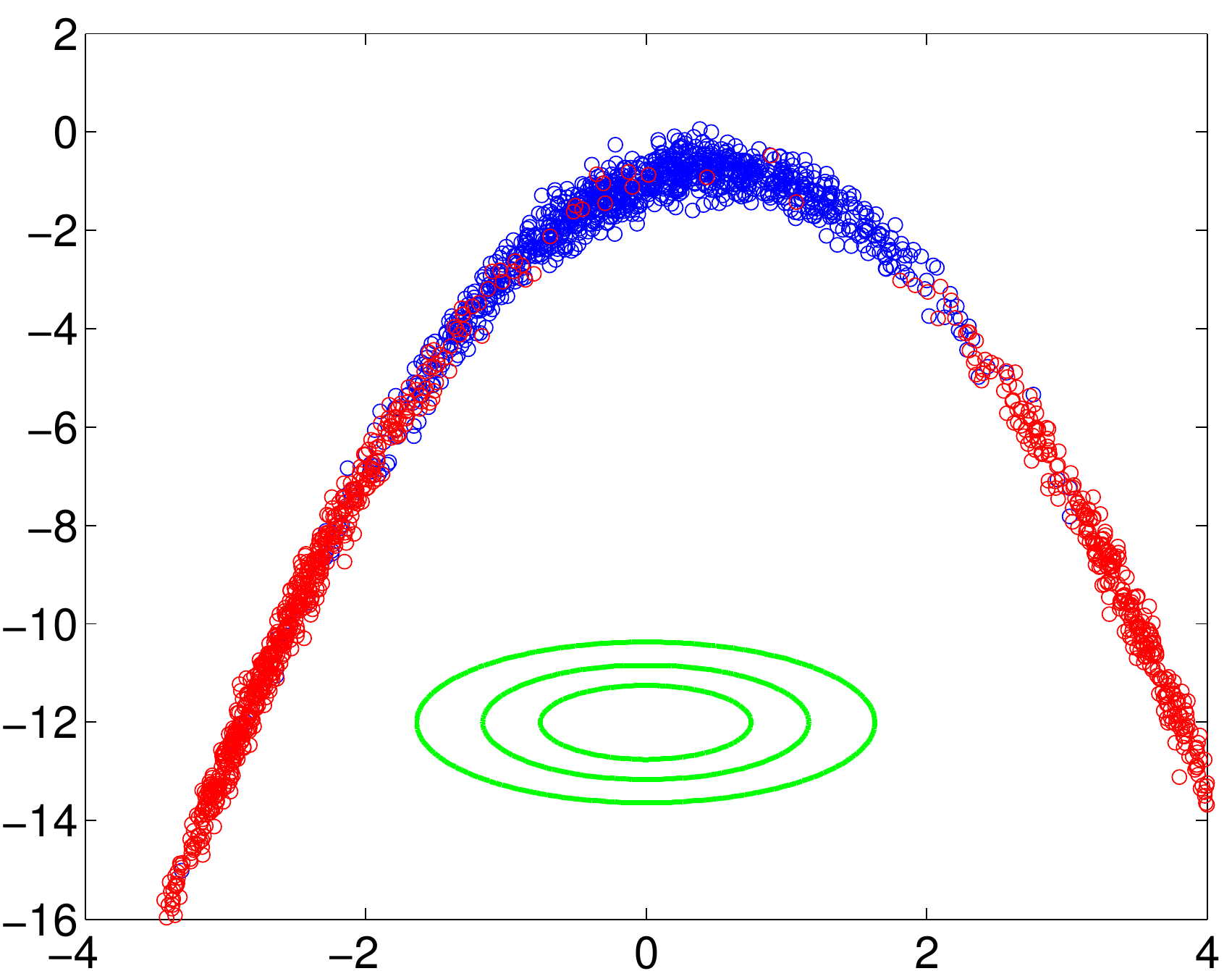}} &
{\includegraphics[scale=0.28]
{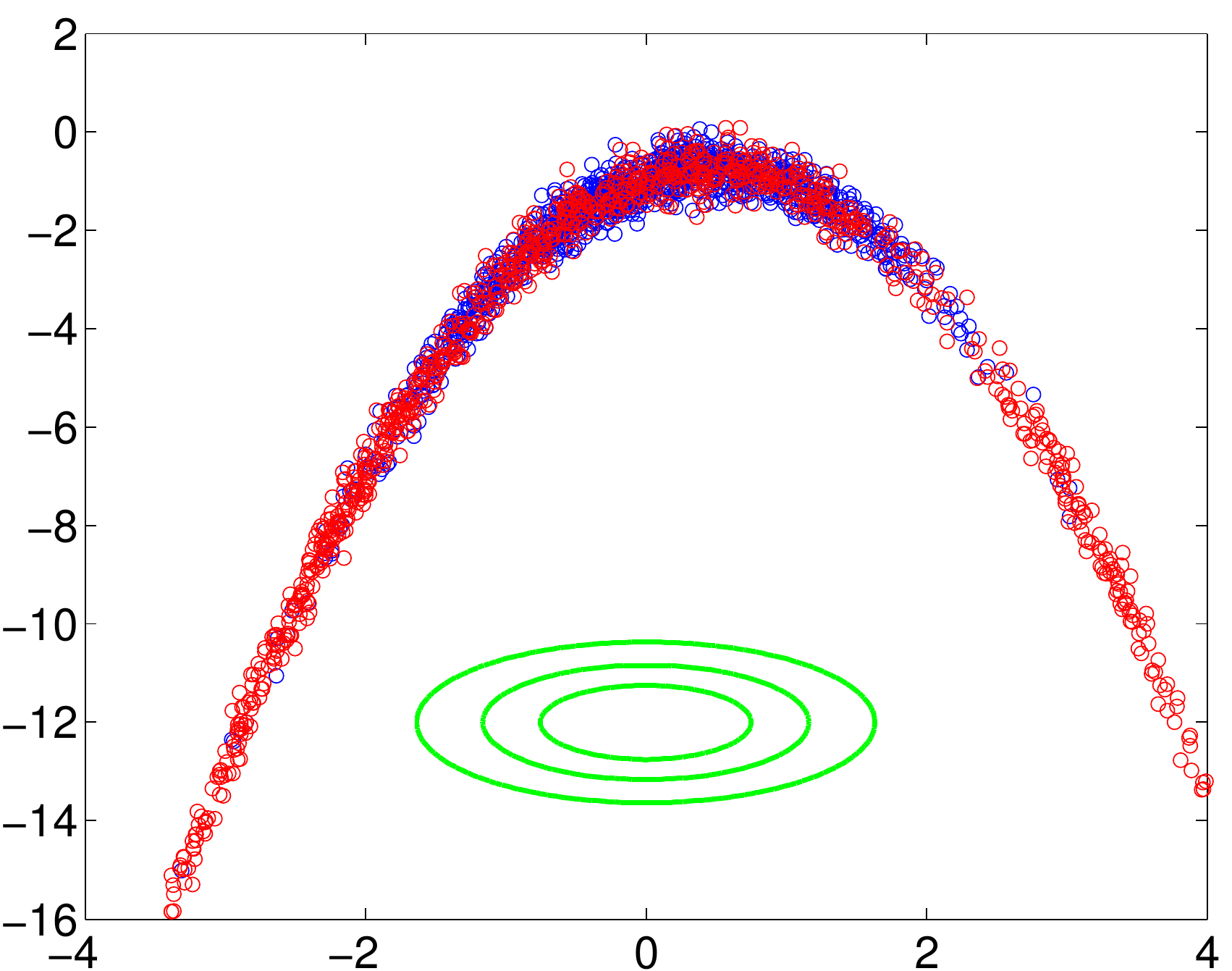}} &
{\includegraphics[scale=0.28]  
{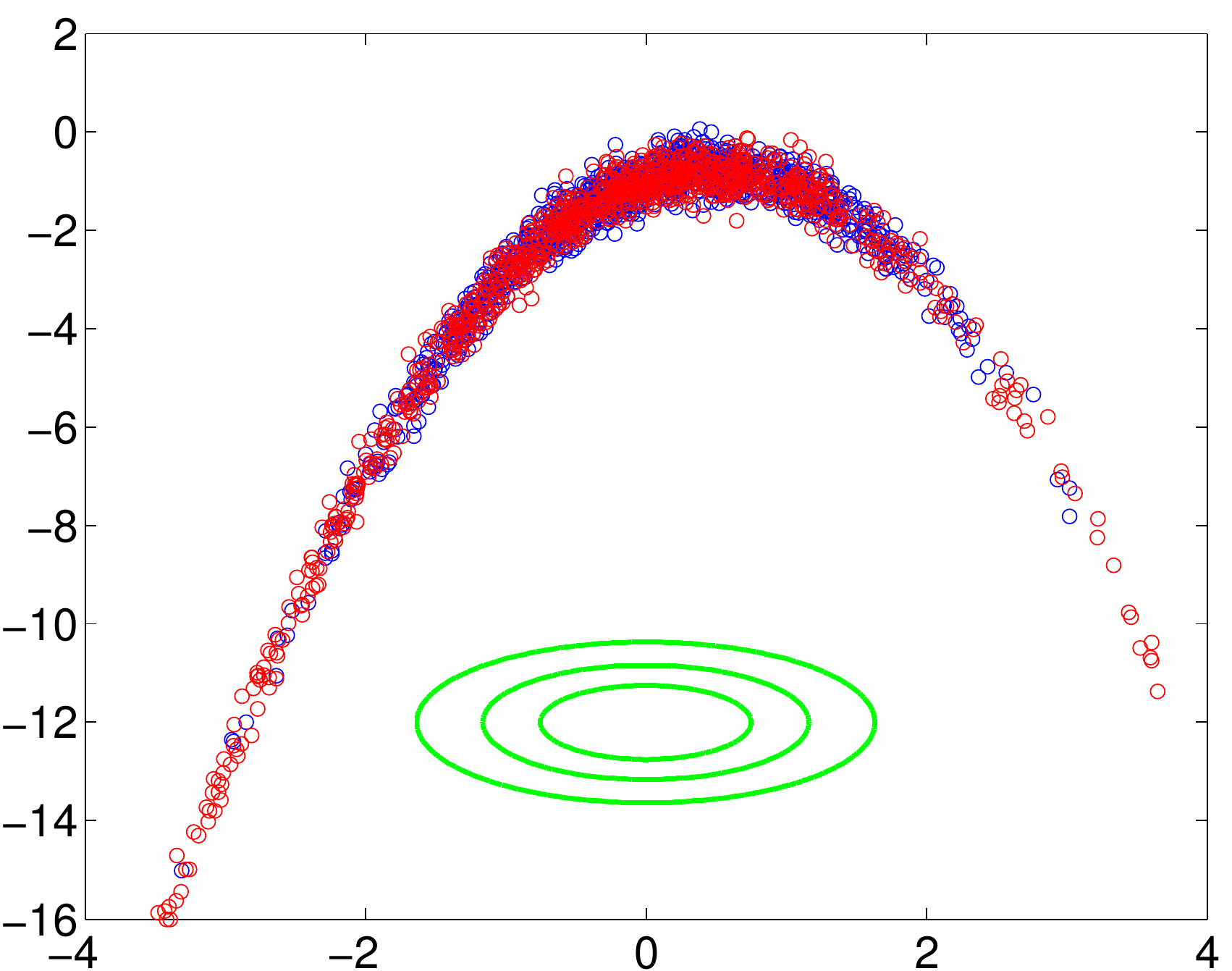}} \\
$t=100$ & $t=500$ &  $t=1000$  
\end{tabular}
\caption{Samples from MCMC marginal distributions (red colour) for different 
number of iterations. Exact samples are shown with blue colour. Green contours show the initial distribution.} 
\label{fig:bananaMH}
\end{figure*}

\begin{figure*}[!htb]
\centering
\begin{tabular}{ccc}
{\includegraphics[scale=0.28]
{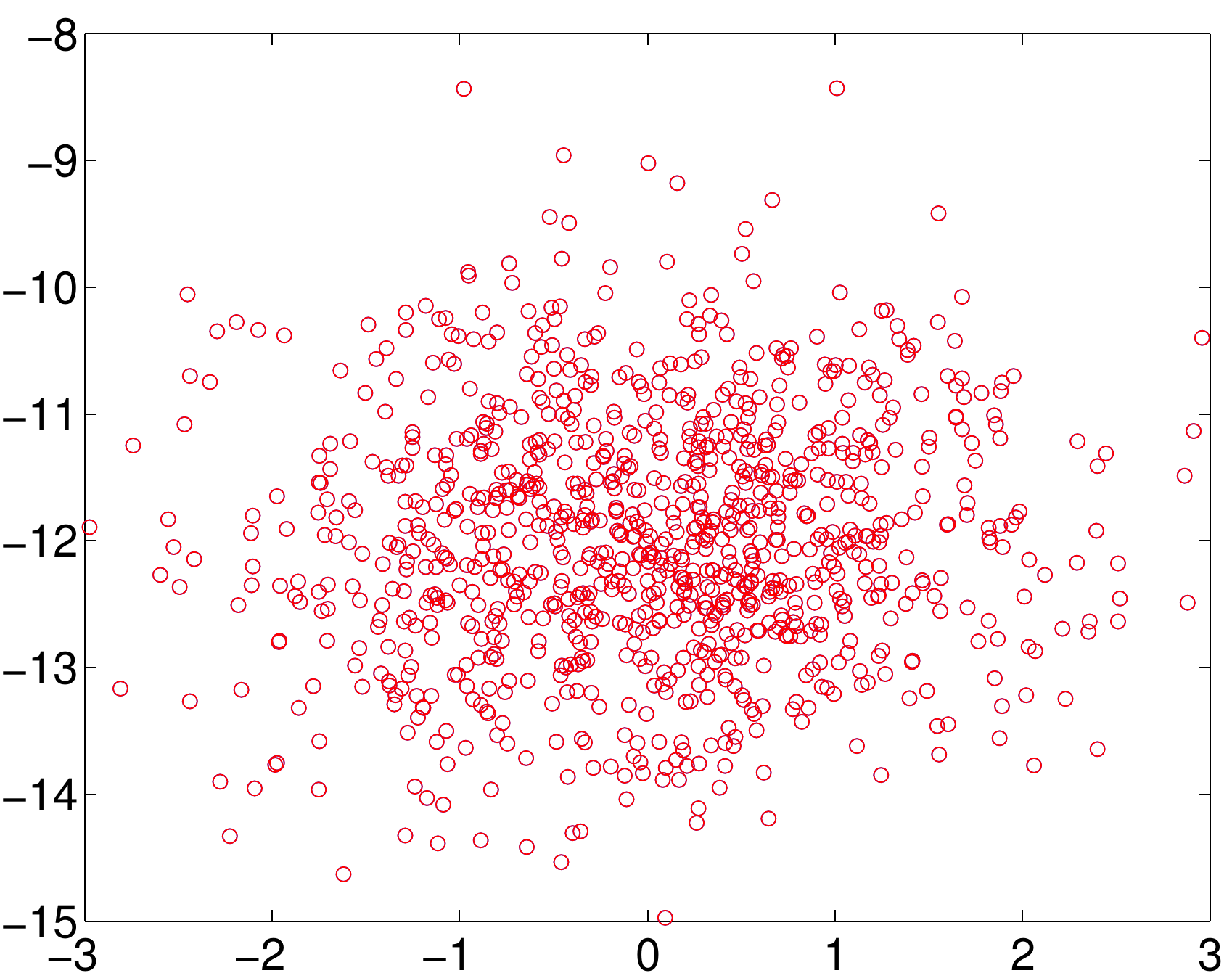}} &
{\includegraphics[scale=0.28]
{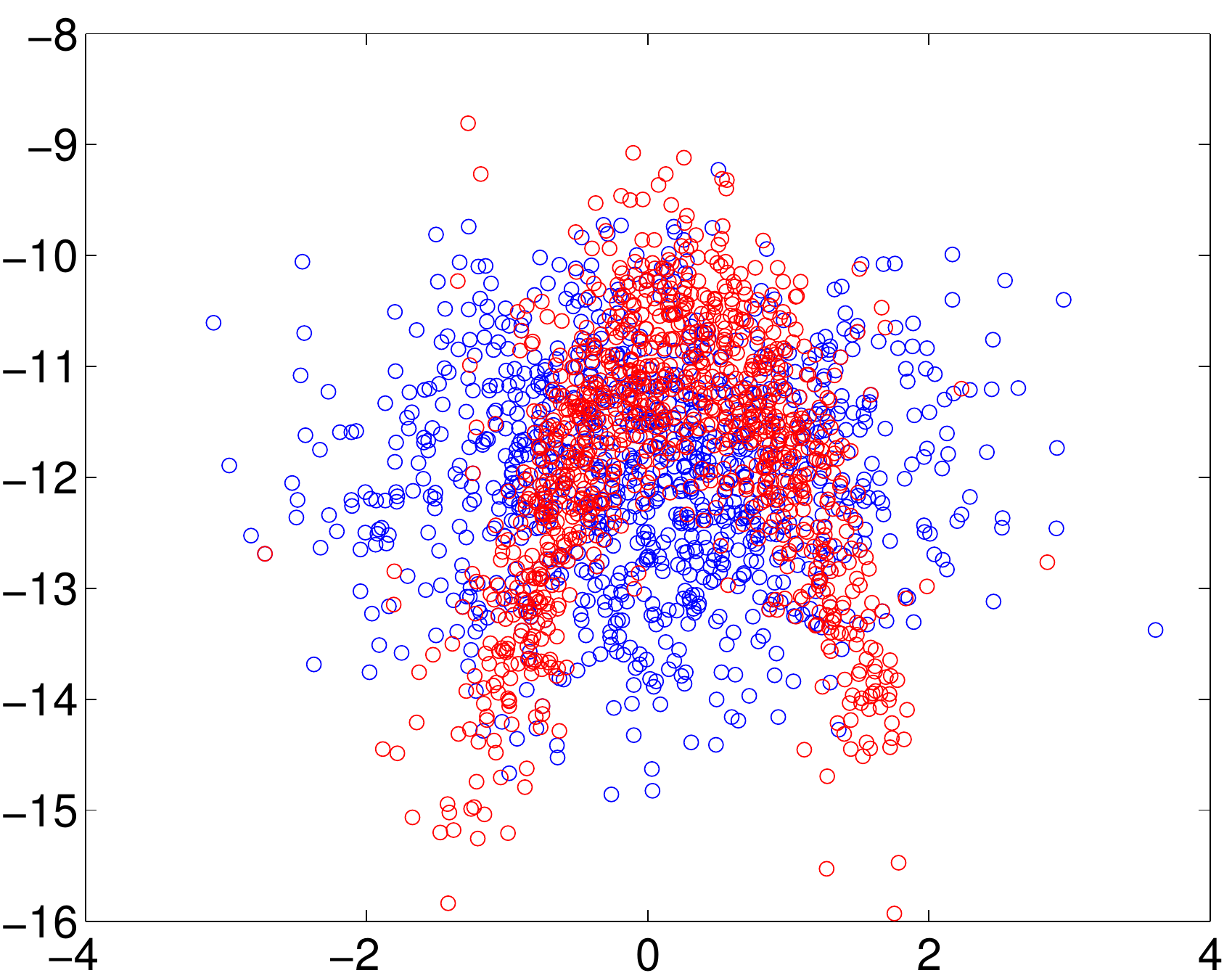}} &
{\includegraphics[scale=0.28]  
{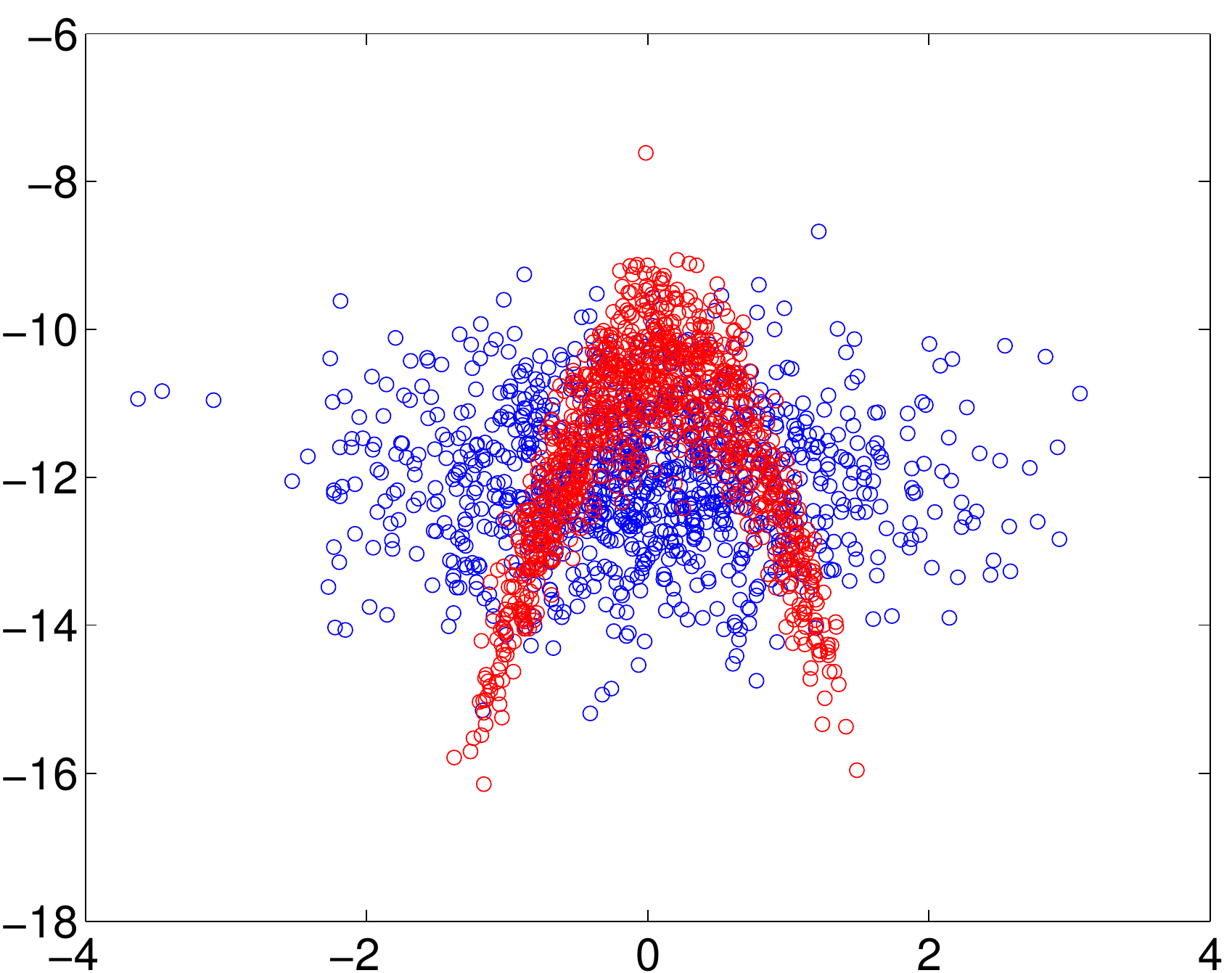}} \\
{\includegraphics[scale=0.28]
{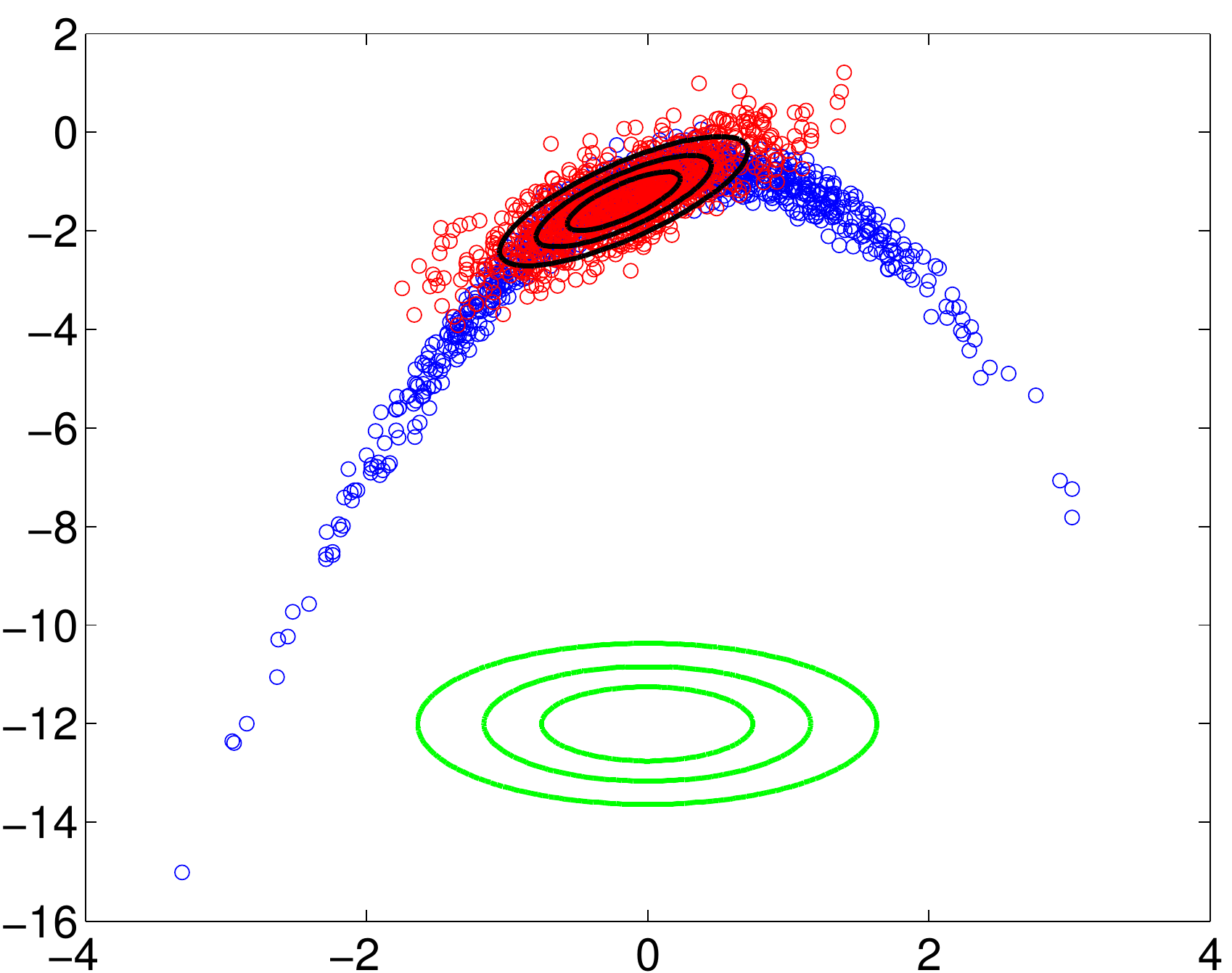}} &
{\includegraphics[scale=0.28]
{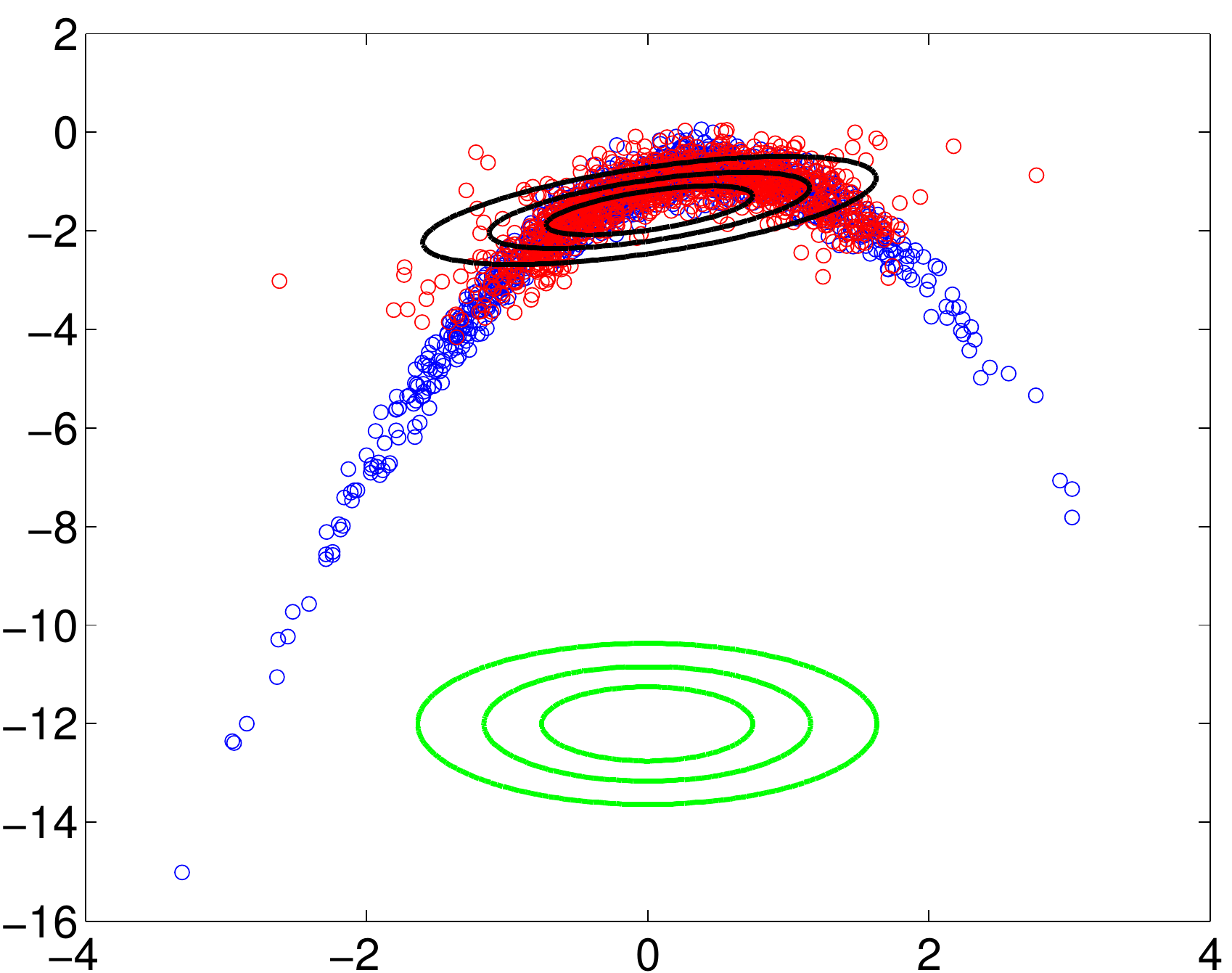}} &
{\includegraphics[scale=0.28]  
{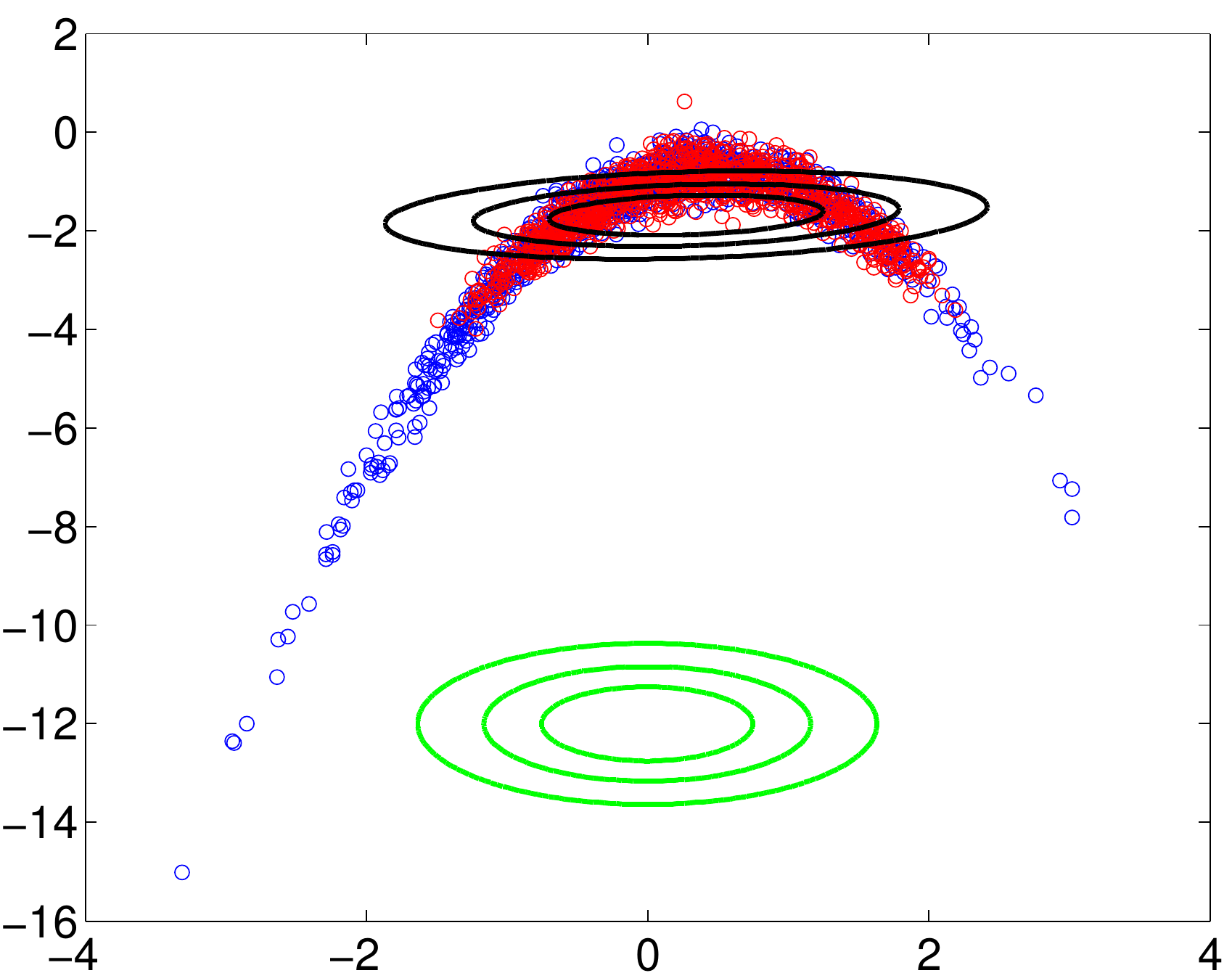}} \\
$t=0$ & $t=5$ & $t=20$ 
\end{tabular}
\caption{Panels in the first row show 
the untransformed samples $\bepsilon_0$ (blue colour) together with the corresponding 
MCMC samples $\bepsilon \sim q_t(\bepsilon)$ (red colour).  
The second row shows exact samples (blue colour) of the banana-shaped distribution 
together with the approximate samples $\bz = L \bepsilon + \bmu$ (red colour). Green contours show the initial distribution and black 
contours visualize the learned transformation parameters $(\bmu,L)$.} 
\label{fig:bananaReparamMH}
\end{figure*}

As an illustrative example we consider the banana-shaped distribution often used to demonstrate 
adaptive MCMC methods \cite{haario-saksman-tamminen-01}. This is defined by taking a two-dimensional 
Gaussian distribution with unit variances and covariance $\rho$ and transform it 
so that the Gaussian coordinates $(\widetilde{z}_1, \widetilde{z}_2)$ become 
$$
z_1 = \alpha \widetilde{z}_1,  \ \ \  z_2 = \frac{\widetilde{z}_2}{\alpha} - b (z_1^2 + \alpha). 
$$
We set $\rho=0.9$ and $\alpha=\beta=1$ which defines the 
distribution with contours shown in Figure \ref{fig:bananaDistr}(a) and exact samples shown  
in Figure \ref{fig:bananaDistr}(b). We first apply MCMC using a  random walk Metropolis-Hastings Gaussian 
proposal $q(\bz'|\bz) = \mathcal{N}(\bz'|\bz, \delta I)$ where the step size $\delta$ 
was set to $\delta=0.5$ that leads to acceptance rates above $40\%$. The initial 
distribution was set to $q_0(\bz_0) = \mathcal{N}(\bz_0| \bmu_0, I)$, with $\bmu_0 = [0 \ \text{-12}]^T$,  depicted
by  the green contours in Figure \ref{fig:bananaMH}. We investigate several 
marginal MCMC distributions $q_t(\bz)$ for different values $t=5,20,250,100,500,1000$. 
Figure \ref{fig:bananaMH} shows independent samples from all these marginals
where we can observe that MCMC converges slowly and even after $1000$ iterations we don't
get an independent sample from the target.  Convergence in this example 
requires few thousands of iterations. Notice that the way we apply 
MCMC in this example is rather unusual since we restart the chain every $t$ iterations 
in order to produce independent samples from a certain marginal $q_t(\bfz)$. An advantage 
of this is that different runs of the chain are trivially parallelizable and when for a certain 
$t$ we have converged all produced samples are truly independent samples from  the target. 
On the other hand, the usual way to run MCMC is based on a single non-parallelizable long chain 
that even when convergence occurs subsequent samples are dependent and could exhibit 
high autocorrelation. 
Also as mentioned in Section \ref{sec:varMCMC} the whole procedure is a special case of Algorithm 1
where $g(\bepsilon,\btheta)$ is chosen to be the identity 
transformation, i.e.\ $\bz = g(\bepsilon,\btheta) = \bepsilon$. 
Next we show that by choosing a different transformation having tunable parameters 
we can significantly improve the approximate inference procedure.    
We consider the affine transformation 
$$
\bz = L \bepsilon + \bmu,   
$$    
where $L$ a is lower triangular positive definite matrix and $\bmu$ a mean vector. 
This transformation is the standard one used in the generic reparametrization-based 
stochastic VI  \cite{Titsias2014_doubly, Kingma2014, Rezende2014} and it can be easily extended to 
account for constraints in the range of values of $\bz$  as done in Stan \cite{Kucukelbir2016}. 
We initialize $L$ to the identity matrix, $\bmu$ to zero vector and 
we use the previous initial distribution $q_0(\bepsilon_0) = \mathcal{N}(\bepsilon_0|\bmu_0, I)$. We then apply Algorithm 1 
for $2000$ iterations, where we learn $(\bmu,L)$, and for  different values $t=0,5,20$. The MCMC transition kernel was based 
on a standard random walk Metropolis-Hastings proposal $\mathcal{N}(\bepsilon'|\bepsilon, \delta I)$. 
The first row of Figure \ref{fig:bananaReparamMH} shows the untransformed initial samples 
$\bepsilon_0 \sim q_0 (\bepsilon_0)$  together with the corresponding MCMC samples 
$\bepsilon \sim q_t(\bepsilon)$.  The second row shows the exact samples 
from the banana-shaped distribution  together with the approximate 
samples $\bz = L \bepsilon + \bmu$, where $\bepsilon \sim q_t(\bepsilon)$. Also the black lines in the second row of  
Figure \ref{fig:bananaReparamMH} visualize the learned 
transformation  parameters $(\bmu, L)$ which are displayed as contours 
of the Gaussian distribution $\mathcal{N}(\bmu + \bmu_0, L L^T)$.  
From Figure \ref{fig:bananaReparamMH} we can observe the following. 
Firstly, for $t=0$ the method reduces to fitting a Gaussian approximation with standard reparametrization-based
stochastic VI. Secondly, as $t$ increases the underlying 
implicit distribution $q_t(\bfz)$ fits better the shape 
of the distribution. The black contours indicate 
that the inferred values $(\bmu, L)$ depend crucially on $t$ (and the underlying MCMC 
operation) and they 
allow to locate the approximation around the highest probability area 
of the banana-shaped distribution. Notice also that due to the nature 
of the KL divergence used the approximation tends 
to be inclusive.

\begin{figure*}[!htb]
\centering
\begin{tabular}{ccc}
{\includegraphics[scale=0.28]
{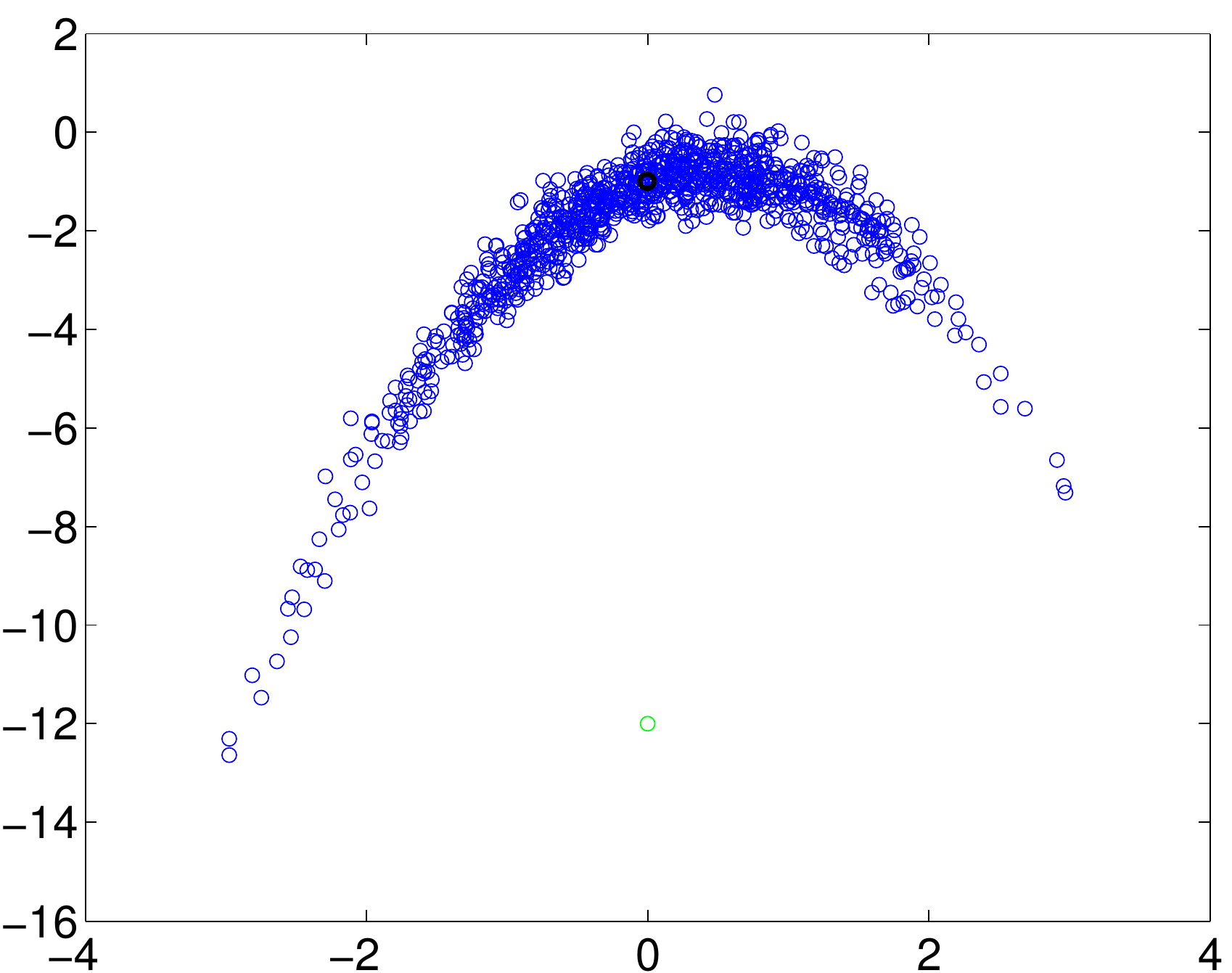}} &
{\includegraphics[scale=0.28]
{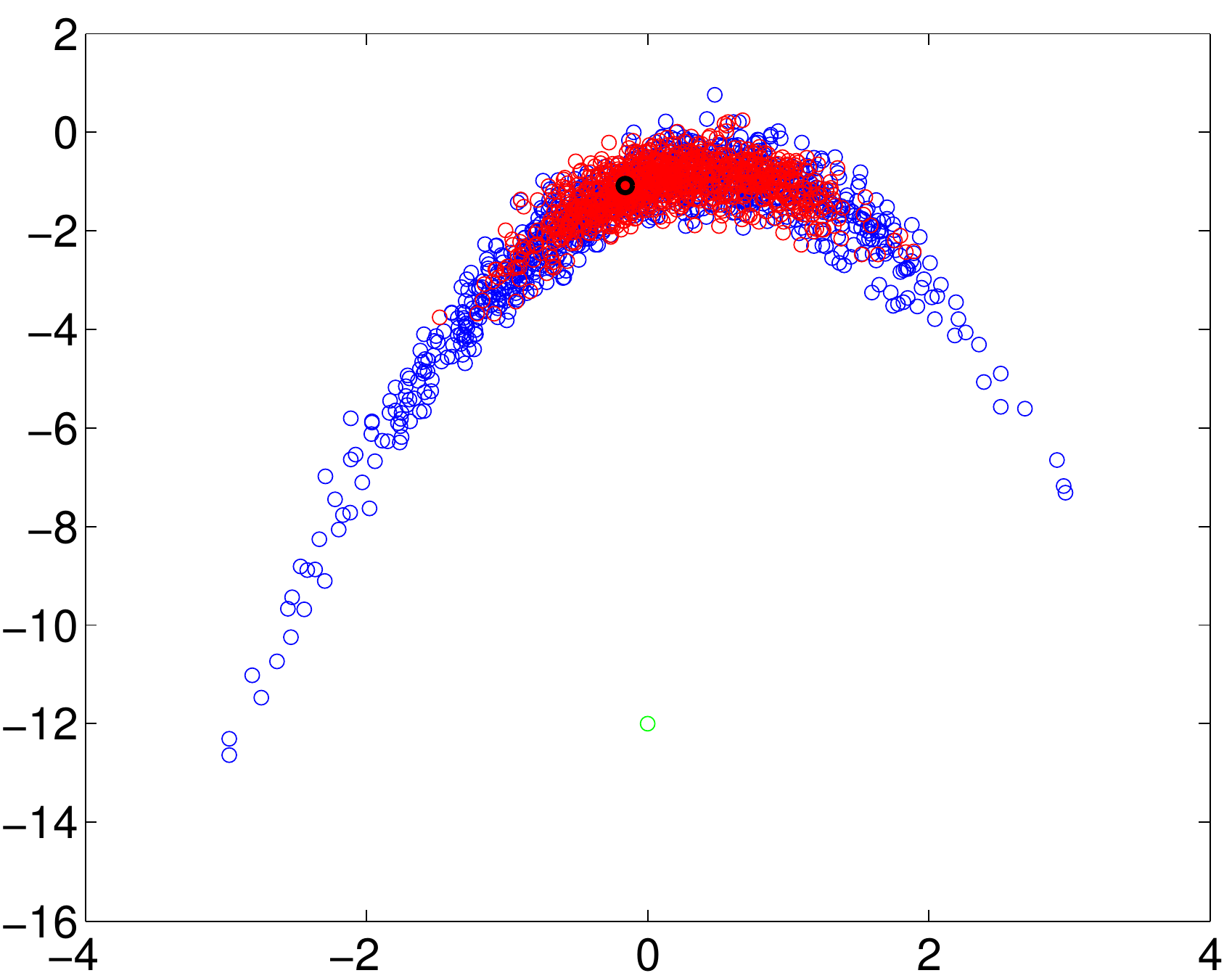}} &
{\includegraphics[scale=0.28]  
{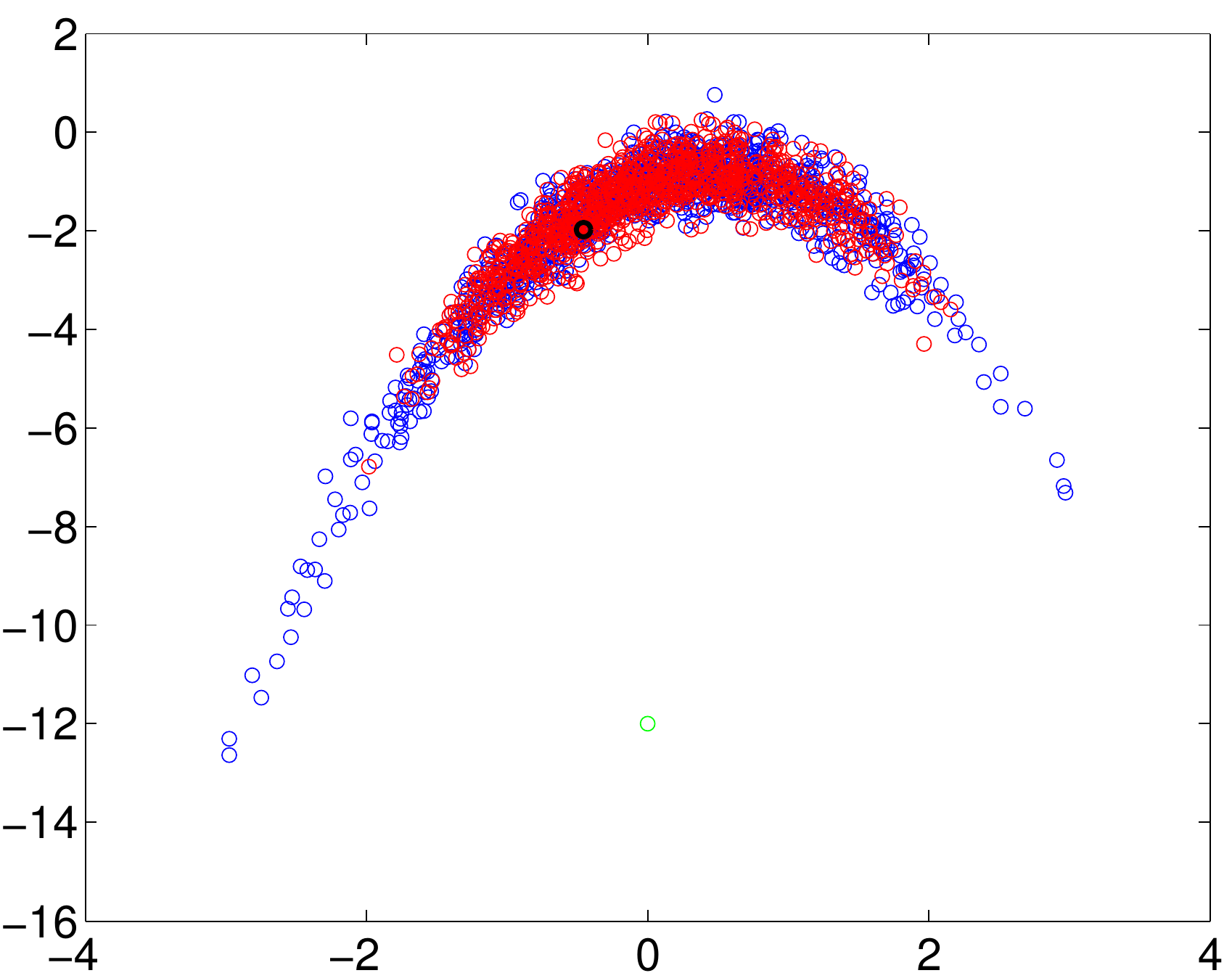}} \\
$t=0$ & $t=5$ & $t=20$ 
\end{tabular}
\caption{Each panel shows exact samples (blue colour) of the banana-shaped distribution 
together with the approximate samples $\bz = \bepsilon + \bmu$ (red colour) 
The green small circles show the initial distribution while 
the black circles show each learned translation vector $\bmu$.} 
\label{fig:bananaReparamMH_Transl}
\end{figure*}

Finally, we repeated all runs for $t=0,5,20$ by simplifying the affine transformation to a simple translation $\bz = \bepsilon + \bmu$  and 
by changing the initial distribution to a point delta function  $q_0(\bepsilon) = \delta(\bepsilon - \bmu_0)$. Figure \ref{fig:bananaReparamMH_Transl}   
 shows the exact samples  from the banana-shaped distribution  together with the approximate 
samples $\bz = \bepsilon + \bmu$. The black small circle shows the learned 
translation vector  $\bmu$ which when $t=0$, as mentioned in Section 2.2, corresponds to the MAP estimate.

\subsection{Variational autoencoders \label{sec:varautoencode}}

Here, we consider an application to amortised inference using variational autoencoders (VAEs) \cite{Kingma2014, Rezende2014}.
Assume a data set $X = \{ \bx_1,\ldots, \bx_n\}$ associated with latent 
variables $Z = \{\bz_1,\ldots, \bz_n\}$ so that each data point $\bx_n$ is generated through a latent variable 
$\bz_n$. The joint distribution of observations and latent variables is given by    
\begin{equation}
p(X,Z) = \prod_{i=1}^m p(\bx_i, \bz_i; \bw).  
\label{eq:jointVAE1}
\end{equation}
In order to speed up training (when $m$ is large) and also prediction in test data we would like to 
amortise inference using a recognition model. Based on the framework in Section \ref{sec:varMCMC} 
we need to define an invertible transformation for each latent variable $\bz_i$, i.e.\
$$
\bz_i = g(\bepsilon_i; \btheta( \bx_i)), \ \ \bepsilon_i =  g^{-1}(\bz_i; \btheta( \bx_i)),
$$
where the vector of reparametrization parameters $\btheta( \bx_i)$ is further parametrized to be a function 
of the observation $\bx_i$. Then from the joint in eq.\ \eqref{eq:jointVAE1} we obtain the reparametrized 
joint given by 
\begin{equation}
p(X,E) = \prod_{i=1}^m p(\bx_i, \bepsilon_i) = \prod_{i=1}^m p(\bx_i, g(\bepsilon_i; \btheta( \bx_i)); \bw ) 
J(\bepsilon_i;  \btheta(\bx_i) ).
\label{eq:jointVAE2}
\end{equation}
The overall variational lower bound is obtained by bounding each log likelihood term $\log p(\bx_i) = \log \int p(\bx_i, \bepsilon_i) d \bepsilon_i$ using a variational distribution $q_i(\bepsilon_i)$ so that
\begin{equation}
\mathcal{F} = \sum_{i=1}^m \int q_i(\bepsilon_i) \left[ 
 \log p(\bx_i, g(\bepsilon_i;\btheta(\bx_i) )) J(\bepsilon_i;  \btheta(\bx_i ))  - \log q_i(\bepsilon_i) \right] d \bepsilon_i. 
\label{eq:boundVAE} 
\end{equation}    
When $q_i(\bepsilon_i)$ is chosen to be the standard normal and $\btheta(\bx_i)$ is an affine transformation of the form 
$$
\btheta(\bx_i) = L(\bx_i) \bepsilon_i + \bmu(\bx_i),
$$
where $L(\bx_i)$ is a positive definite diagonal matrix and $\bmu(\bx_i)$ a real-valued mean vector 
(typically both parametrized by neural nets), then the maximization of the lower bound in \eqref{eq:boundVAE} reduces to the standard  
variational autoencoder training procedure with factorised Gaussian approximations \cite{Kingma2014, Rezende2014}. Next, we compare this 
standard approach with our proposed MCMC-based variational approximation where 
we define each $q_i(\bepsilon_i)$ to be a $t$-step MCMC distribution.  
   
\begin{figure*}[!htb]
\centering
\begin{tabular}{cc}
{\includegraphics[scale=0.4]
{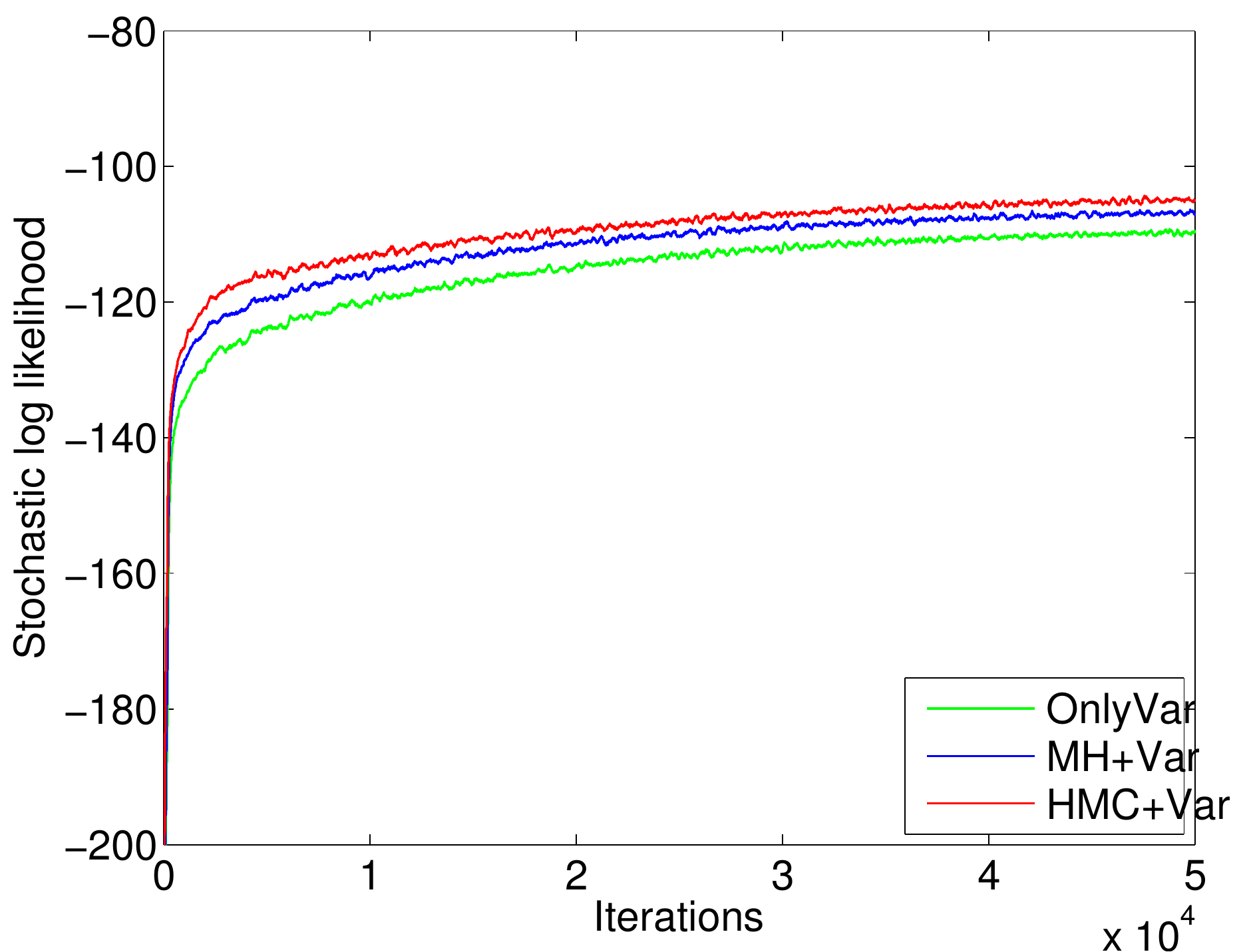}} &  
{\includegraphics[scale=0.4]
{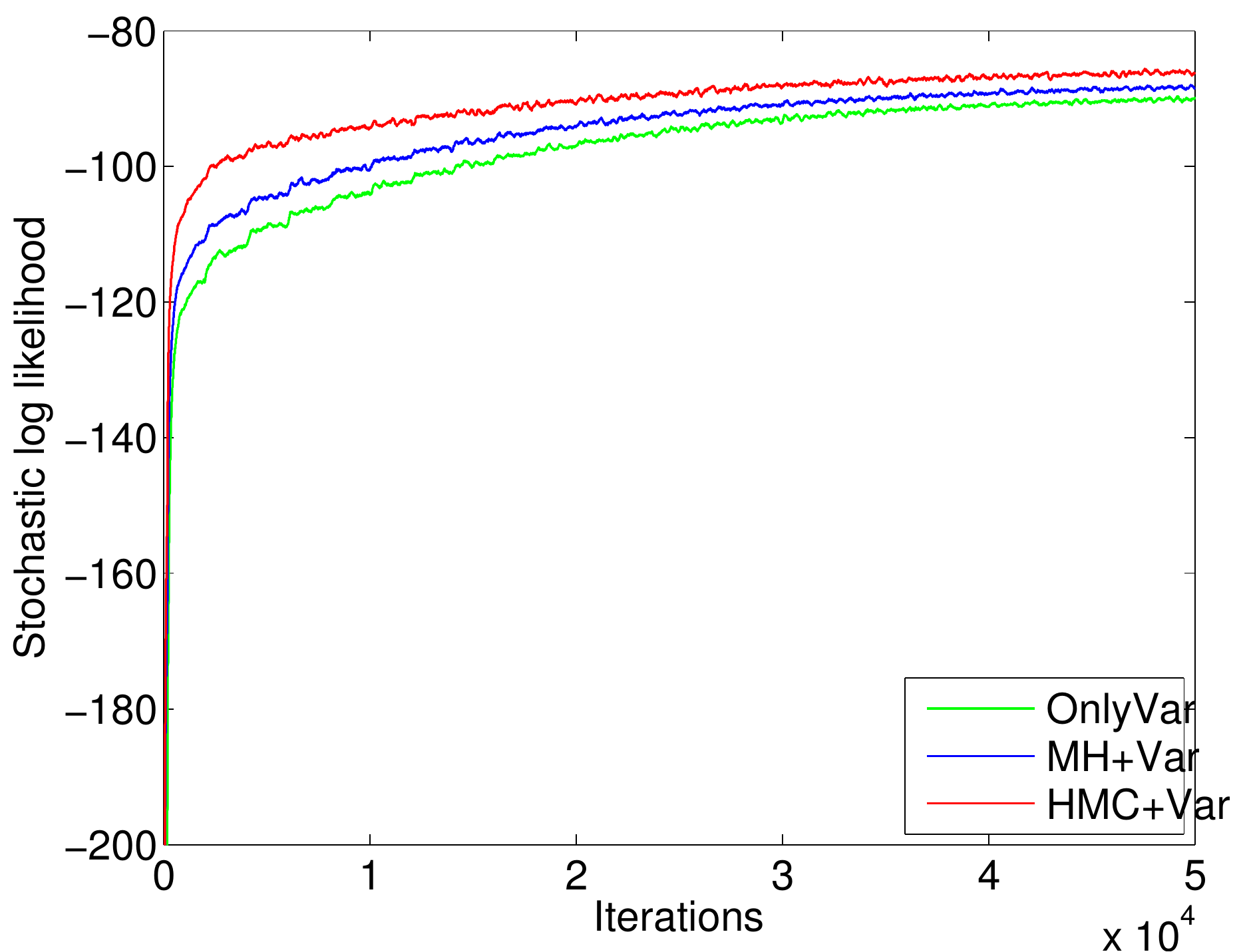}}  \\
$n=5$  & $n=10$
\end{tabular}
\caption{The left panel shows the stochastic bounds (smoothed using a rolling window over $200$ actual values) without the $ - \log q_i(\bz_i)$ term 
for the three compared methods when the  dimensionality of the variable $\bz_i$ is $n=5$, while the right panel 
shows the same quantities for $n=10$. The label OnlyVar indicates the standard VAE with a Gaussian approximation, while
HMC$+$Var  and MH$+$Var indicate two proposed  schemes that combine MCMC and VI.} 
\label{fig:mnistbounds}
\end{figure*}

We consider the $50000$ binarized MNIST digits and construct the latent variable model 
$p(\bx_i, \bz_i; \bw)$ based on a single hidden layer neural net having as input $\bz_i$
and sigmoid outputs that model the probabilities of the factorised Bernoulli  
likelihood $p(\bx_i| \bz_i; \bw)$. Each prior $p(\bz_i)$ over the latent variables was chosen to be standard 
normal. This defines a latent variable model for binary data 
where the essential difference with traditional models, such binary PCA and factor analysis,
is that the mapping from $\bz_i$ to $\bx_i$ is non-linear and it is parametrized by a neural net. 
For the hidden layer we use $200$ relu units. The diagonal non-negative elements of $L(\bx_i)$ and the real-valued vector 
$\bmu(\bx_i)$ were also parametrized with separate neural nets each having a single hidden layer
with $200$ relu units. We compare three VAE training procedues: i) the standard VAE using 
Gaussian approximations where each $q_i(\bepsilon_i)$ is a standard normal, ii) a variational approximation 
corresponding to an MCMC marginal distribution obtained by initializing with a standard normal and then running $t=2$ Hamiltonian Monte Carlo (HMC) iterations with $5$ leap-frog steps and ii) and similarly by using a MCMC marginal obtained by initializing with a standard normal and then running $t=10$ random walk Metropolis-Hastings (MH) steps with a Gaussian proposal distribution. 
For the parameters $\bw$ of the latent variable model we carry out point estimation (see Algorithm 1).  
All methods were applied using the same learning schedule for $5 \times 10^4$ iterations 
with minibatch size $100$ and by using only one sample over $\bepsilon_i$ when doing parameter updates. The step sizes of  
HMC and MH were tuned so that to obtain acceptance rates around $90\%$ and $40\%$ respectively. 
Figure \ref{fig:mnistbounds} displays the evolution of the stochastic bound without the intractable  
$ -\log q_i(\bepsilon)$ term\footnote{Recall that for the proposed MCMC-based variational method $\log q_i(\bepsilon)$ is not tractable since $q_i(\bepsilon)$ is implicit.} (i.e.\ only the  log joint term that indicates data reconstruction)
for all three methods. Clearly, both MCMC-based methods lead to better 
training since they allow to express more flexible implicit variational distributions that can fit better the shape of
each latent variable posterior. Also the HMC-based marginal works better that the MH-based marginal. 

\begin{figure*}[!htb]
\centering
\begin{tabular}{c}
{\includegraphics[scale=0.8]
{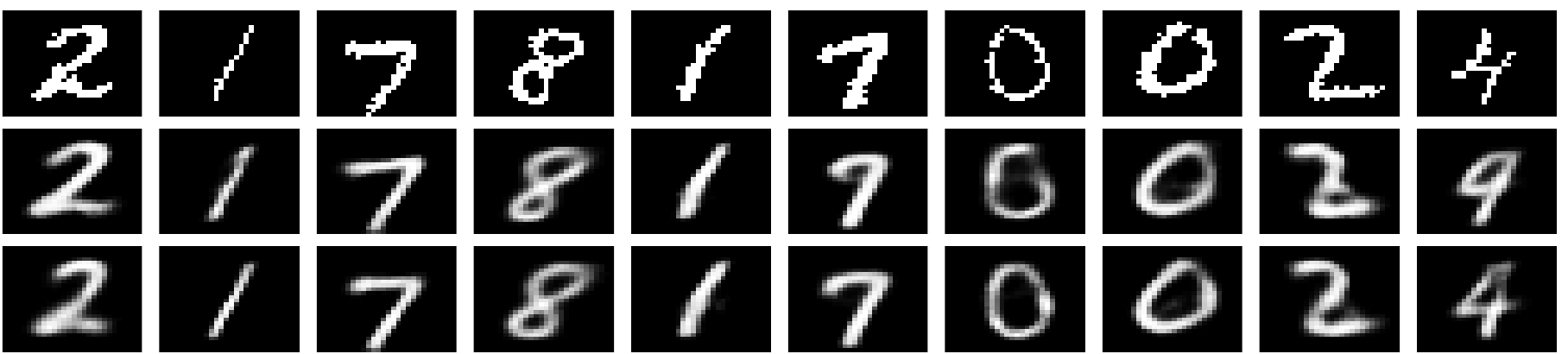}} 
\end{tabular}
\caption{The panels in the first row show some test digits. The panels in the 
second row show reconstructions obtained by the standard Gaussian approximation (i.e.\ the OnlyVar method)
while the panels in the third row show the reconstructions of the HMC+Var method. These reconstructions 
were obtained by assuming latent dimensionality $n=5$. The average reconstructions across all $10000$ test examples, 
measured by cross entropy between each binary digit and its predictive probability vector, was $-111.2939$ for OnlyVar and 
$-104.1400$ for HMC+Var. When $n=10$ the corresponding numbers were  $-87.3213$ and $-82.3134$.} 
\label{fig:mnistReconstruct}
\end{figure*}

Amortised inference allows to quickly express an variational approximation to some latent variable posterior 
$p(\bz_*| \bx_*; \bw)$, where for instance $\bx_*$ could be a test example, without requiring to perform further 
optimization. For the proposed methods this approximation is implicit, i.e.\ 
we first express $\bepsilon_* \sim q_*(\bepsilon_*)$ by running $t$ MCMC iterations and then 
an independent sample from the approximate posterior is given by $\bz_* = L(\bx_*) \bepsilon_* + \bmu(\bx_*)$. 
Such a sample $\bz_*$ can be thought of as a possible latent representation of $\bx_*$ that can be used for instance 
to reconstruct $\bx_*$. Figure \ref{fig:mnistReconstruct} shows some test data (first row) together with the 
corresponding reconstructions obtained by the standard Gaussian approximation (second row) and by the 
HMC-based variational method (third row). Clearly, the reconstructions based on the proposed 
method are better. For instance, the right most digit is reconstructed as "9" by the Gaussian approximation
while  correctly is reconstructed as "4" by the proposed method. Further details and quantitative  
scores across all $10000$ test data are given in the caption of Figure \ref{fig:mnistReconstruct}.

%

\section{Discussion  \label{sec:discussion} }

We have presented an algorithm that combines MCMC and variational inference  and it can 
auto-tune its performance  by learning model-based reparametrizations. 
This is an implicit variational inference algorithm that makes use of the reparametrization trick and  
 fits complex MCMC distributions to exact posteriors without requiring the estimation of log density ratios. 

A limitation of the current approach is that it is applicable only to models where we can define a learnable model-based 
reparametrization (this is typically possible for continuous and differentiable models). For the future it would 
 be interesting to develop alternative and more general ways to combine MCMC and variational inference that would be applicable 
 to arbitrary models such as those that require inference over discrete variables. A preliminary attempt to 
 develop a general procedure based on the score function method, that covers also discrete variables, 
  is described in the Appendix.


\begin{small}
\subsubsection*{References} 
\renewcommand{\section}[2]{}%
\bibliographystyle{plain}
\bibliography{ref,bibReparamGrad}
\end{small}

\appendix

\section{MCMC-based implicit variational inference using the score function method}

The term explicit distribution refers to a distribution 
such that we can both simulate from and tractably evaluate its density, 
while the term implicit distribution refers to a distribution from which we can only sample. 
Next the Dirac delta mass  $\delta(\bz)$  would be considered as an explicit distribution. 
A general way to construct implicit distributions 
is by using the following mixture representation
$$
q(\bz) = \int Q(\bz|\bz_0) q(\bz_0) d \bz_0,
$$    
where $\bz_0$ does not have to live in the same space as $\bz$. The simplest choice 
of having an implicit $q(\bz)$ is to assume that both $q(\bz_0)$ and $Q(\bz|\bz_0)$ 
are explicit but the integral is intractable. An example of this, 
very popular in recent  complex generative models based on adversarial
training \cite{GAN2014}, is to set   $q(\bz_0)$ to be a simple Gaussian or uniform, 
and define  $Q(\bz|\bz_0)$ from a neural net mapping $\bz = f(\bz_0;\btheta)$ having 
as input the vector $\bz_0$ and parameters $\btheta$. The latter essentially sets  
$Q(\bz|\bz_0)$ equal to the Dirac delta function $\delta(\bz - f(\bz_0;\btheta))$. 
Other ways to obtain an implicit $q(\bz)$ is by allowing $q(\bz_0)$ and/or 
$Q(\bz|\bz_0)$ to be also implicit. The MCMC-based variational inference method 
that we presented in the main paper belongs to this latter category where 
$q(\bz_0) \equiv q_t(\bepsilon)$ is an implicit MCMC marginal,  
$Q(\bz|\bz_0) \equiv \delta(\bz - g(\bepsilon;\btheta))$ and $ \bz_0 \equiv \bepsilon$. The ability 
to reparametrize the variational lower bound was due to the invertibility of $g(\bepsilon;\btheta)$. 
Next we will discuss a different combination of MCMC and variational inference 
by setting $q(\bz_0)$ to a simple explicit distribution and constructing  
$Q(\bz|\bz_0)$ based on MCMC. 

Suppose a MCMC marginal distribution $q_t(\bz)$ with transition kernel $q(\bz'|\bz)$ 
that targets the posterior of interest $p(\bz|\bx)$. Specifically, $q_t(\bz)$ is constructed by an 
initial distribution $q_0(\bz_0;\btheta)$, that depends on tunable 
parameters $\btheta$, and the compound transition density $Q_t(\bz|\bz_0)$ corresponding to 
$t$ MCMC iterations, i.e.\ 
\begin{align}
q_t(\bz)  = \int Q_t(\bz|\bz_0) q_0(\bz_0; \btheta) d \bz_0.
\label{eq:qtz2}
\end{align} 
Notice that $Q_t(\bz|\bz_0)$ is implicit while $q_0(\bz_0; \btheta)$ is explicit. 
We wish to use $q_t(\bz_t)$  as a variational distribution and optimize it over the 
parameters $\btheta$ that determine the explicit component $q_0(\bz_0; \btheta)$. 
Since $q_0(\bz_0; \btheta)$ determines the initialization of MCMC, by tuning $\btheta$
we are essentially learning how to start in order to speed up MCMC convergence within the budget of $t$
iterations.  The minimization of $\text{KL}(q_t(\bz) || p(\bz|\bx))$ leads to the maximization of  
the lower bound 
\begin{equation}
\mathcal{F}_t(\btheta) = \mathbbm{E}_{q_t(\bz)}[\log p(\bx,\bz) - \log q_t(\bz)]. \nonumber 
\label{eq:lowerqt}
\end{equation} 
By taking gradients  with respect to $\btheta$ and by making use of the identity 
$\int q_t(\bz) \nabla_{\btheta} \log q_t(\bz) d \bz = 0$  we have 
\begin{align}
\nabla_{\btheta} \mathcal{F}_t(\btheta) & = \int_{\bz}  \nabla_{\btheta} q_t(\bz) \left[ \log p(\bx,\bz) - \log q_t(\bz)
\right] d \bz \nonumber \\ 
& = \int_{\bz}  \left( \int_{\bz_0} Q_t(\bz |\bz_0)  \nabla_{\btheta} q_0(\bz_0; \btheta) d \bz_0 \right) 
 \left[ \log p(\bx,\bz) - \log q_t(\bz)
\right]  d \bz \nonumber \\ 
& = \int_{\bz}  Q_t(\bz |\bz_0)  \left( \int_{\bz_0}  q_0(\bz_0; \btheta)  \nabla_{\btheta} \log q_0(\bz_0; \btheta) d \bz_0 \right)  \left[ \log p(\bx,\bz) - \log q_t(\bz)
\right]  d \bz, \nonumber 
\end{align}
where in the second line we used \eqref{eq:qtz2} and in third one the score 
function method, also called log-derivative trick or
REINFORCE \cite{Williams1992,Glynn1990}. An unbiased estimate of the gradient can be obtained by drawing 
a set of samples $\{\bz_0^{(s)}, \bz^{(s)}\}_{s=1}^S$ where each $\bz_0^{(s)} \sim q_0(\bz_0; \btheta)$ 
 and $\bz^{(s)} \sim Q_t(\bz|\bz_0^{(s)})$, i.e.\ $\bz^{(s)}$ is the final  state of an MCMC run of length $t$ 
when starting from $\bz_0^{(s)}$, and then using   
\begin{equation}
\frac{1}{S} \sum_{s=1}^S \nabla_{\btheta} \log q_0(\bz_0^{(s)}; \btheta) \left[ \log p(\bx,\bz^{(s)}) - 
\log q_t(\bz^{(s)}) \right]. 
\end{equation}
This estimate is still intractable because $q_t(\bz)$ is implicit and we cannot evaluate the term 
$\log q_t(\bz)$.  A way to get around this problem is to approximate $\log q_t(\bz)$ by applying 
log density ratio estimation. However, this can be very difficult in high-dimensional settings and therefore the ability to get reliable
estimates for $\log q_t(\bz)$ is the main obstacle for the general applicability  of the above framework. 


\end{document}